\documentclass[runningheads,a4paper]{llncs}

\usepackage{amssymb}
\usepackage{stmaryrd}
\usepackage{makeidx}  
\usepackage[lined,ruled,vlined,algoruled]{algorithm2e}

\begin{document}
\mainmatter              
\title{Algorithm for Adapting Cases Represented in a Tractable Description Logic}
\titlerunning{Adapting Cases Represented in a Tractable Description Logic}

\author{Liang Chang\inst{1,2} \and Uli Sattler \inst{2} \and Tianlong Gu \inst{1}}
\institute{
Guangxi Key Laboratory of Trusted Software,
\\Guilin University of Electronic Technology,
Guilin, 541004, China
\\
\and
School of Computer Science,
\\The University of Manchester,
Manchester, M13 9PL, UK
\\
\email{changl.guet@gmail.com,sattler@cs.manchester.ac.uk,cctlgu@guet.edu.cn}}
\maketitle              

\begin{abstract}  
Case-based reasoning (CBR) based on description logics (DLs) has gained a lot of attention lately.
Adaptation is a basic task in CBR that can be modeled as a knowledge base revision problem
which has been solved in propositional logic.
However, in DLs, adaptation is still a challenge problem since existing revision operators only work well for
DLs of the \emph{DL-Lite} family.
It is difficult to design revision algorithms that are syntax-independent and fine-grained.
In this paper,
we present a new method for adaptation based on the tractable DL $\mathcal{EL_{\bot}}$.
Following the idea of adaptation as revision,
we firstly extend the logical basis for describing cases from propositional logic to the DL $\mathcal{EL_{\bot}}$,
and then present a formalism for adaptation based on $\mathcal{EL_{\bot}}$.
With this formalism,
we show that existing revision operators and algorithms in DLs do not work for it,
and then present our adaptation algorithm.
Our algorithm is syntax-independent and fine-grained, and satisfies the requirements on revision operators.
\end{abstract}

\section{Introduction}   \label{section:introduction}

Description logic (DL) is a family of logics for representing and reasoning about
knowledge of static application domains \cite{baader:03}.
It is playing a central role in the Semantic Web,
serving as the basis of the W3C-recommended Web ontology language OWL \cite{horrocks:03}.
The main strength of DLs is that they offer considerable expressive power often going far
beyond propositional logic, while reasoning is still decidable.
Furthermore, DLs have well-defined semantics and are supported by many
efficient reasoners.


In the last few years, there has been a growing interest in bringing the power and character of
DLs into case-based reasoning (CBR) \cite{d'aquin:05,gomez:99,salotti:98,sanchez:11}.
CBR is a type of analogical reasoning in which new problem is solved by reusing past experiences called \emph{source cases}.
%
%
%
%
There are two basic tasks in the CBR inference: retrieval and adaptation.
\emph{Retrieval} aims at selecting a source case that is similar to the new problem according to some similarity criterion.
\emph{Adaptation} aims at generating a solution for the new problem by adapting the solution contained in the source case.
At present, most research is concerned with the
retrieval task when introducing DLs into CBR
\cite{d'aquin:05,salotti:98,sanchez:11}.

In comparison to retrieval, adaptation is often considered to be the more difficult task.
One approach for this task is to model the adaptation process as the \emph{revision} problem of a knowledge base (KB) \cite{lieber:07,cojan:10};
it is hoped that an adaptation algorithm satisfies the AGM postulates on revision operators \cite{alchourron:85,katsuno:91}.
In propositional logic, there are many revision operators
which satisfy the AGM postulates and can be applied to complete the adaptation task \cite{lieber:07}.
However, in DLs, it is very difficult to design revision operators and algorithms
that satisfy the AGM postulates \cite{flouris:05}.
Especially, it is a great challenge to design revision algorithms that are independent of the syntactical forms of KBs
and fine-grained for the minimal change principle.

There are two kinds of revision operators and algorithms in the literature:
\emph{model-based approaches} (MBAs) and \emph{formula-based approaches} (FBAs).
Revision operators of MBAs can be treated as DL-based extensions of the classical revision operators
in propositional logic,
in that the semantics of minimal change is defined by measuring the
distance between models \cite{kharlamov:13}.
MBAs are syntax-independent and fine-grained,
but at present they only work for DLs of the \emph{DL-Lite} family.
In FBAs, the semantics of minimal change is reflected in the minimality of formulas removed by the revision process.
There are two FBAs in the literature. One is based on the deductive closure of a KB \cite{calvanese:10,lenzerini:11};
it is syntax-independent and fine-grained, but again only works for DLs of the \emph{DL-Lite} family.
Another is based on justifications \cite{wiener:06};
although it is applicable to DLs such as $\mathcal{SHOIN}$,
it is syntax-dependent and not fine-grained.

%
%


DLs of the $\mathcal{EL}$ family are popular for building large-scale ontologies \cite{baader:05b}.
Some important medical ontologies and life science ontologies are built in $\mathcal{EL}$,
such as the SNOMED CT \cite{spackman:00}
and the Gene Ontology \cite{consortium:00}.
A feature of this family of DLs is that they allow for reasoning in polynomial time,
while being able to describe ``relational structures''.
They are promising DLs for CBR since they are,
on the one hand, of interesting expressive power (orthogonal to \emph{DL-Lite}) and,
on the other hand, restricted enough so that we can hope for a practical adaptation approach.
In the literature, some good results on introducing
DLs of the $\mathcal{EL}$ family into the retrieval of source cases
have been presented \cite{sanchez:11};
the problem of measuring the similarity of concepts in these DLs is also well-studied
\cite{lehmann:12}.
However, adaptation based on these DLs is still an open problem.
The reason is that existing revision operators, to the best of our knowledge,
are not syntax-independent and fine-grained.



In this paper we present a new method for adaptation in the DL $\mathcal{EL_{\bot}}$ of the $\mathcal{EL}$ family.
Our contributions regard three aspects.
Firstly, we extend the logical basis for describing cases from propositional logic to the DL $\mathcal{EL_{\bot}}$,
with a powerful way of describing cases as ABoxes in DL.
Secondly, we extend the ``adaptation as KB revision" view from \cite{lieber:07}
to the above setting and get a formalism for adaptation based on $\mathcal{EL_{\bot}}$.
Finally, for the adaptation setting we provide an adaptation algorithm.
Our algorithm is syntax-independent and fine-grained,
and satisfies the requirements on revision operators.

\section{The Description Logic $\mathcal{EL_{\bot}}$} \label{section:DL}

The DL $\mathcal{EL_{\bot}}$ extends $\mathcal{EL}$ with bottom concept (and consequently
disjointness statements) \cite{baader:05b}.
Let $N_{C}$, $N_{R}$ and $N_{I}$ be disjoint sets of
\emph{concept names}, \emph{role names} and \emph{individual names}, respectively.
$\mathcal{EL_{\bot}}$-\emph{concepts} are built according to the following syntax rule
$  C  ::=  \top \:|\: \bot \:|\: A \:|\: C\sqcap D \:|\: \exists r.C$,
where $A \in N_C$, $r \in N_R$, and $C, D$ range over $\mathcal{EL_{\bot}}$-concepts.

A \emph{TBox} $\mathcal{T}$ is a finite set of \emph{general concept inclusions} (GCIs)
of the form $C \sqsubseteq D$, where $C$ and $D$ are concepts.
An \emph{ABox} $\mathcal{A}$ is a finite set of \emph{concept assertions} of
the form $C(a)$ and \emph{role assertions} of the form $r(a,b)$, where $a,b \in N_{I}$, $r \in N_{R}$, and $C$ is a concept.
A \emph{knowledge base} (KB) is a pair $\mathcal{K}=\langle \mathcal{T},\mathcal{A} \rangle$.

\begin{example} \label{example:cancer treatment}
Consider the example on breast cancer treatment discussed in \cite{lieber:07}.
We add some background knowledge to it and describe the knowledge by the following GCIs
in a TBox $\mathcal{T}$:
\begin{eqnarray*}
   & & Tamoxifen \sqsubseteq Anti\textrm{-}oestrogen,
   \;\;\;\;\;\;\;\;\;\;
   Anti\textrm{-}aromatases \sqsubseteq Anti\textrm{-}oestrogen,
   \\
   & & Tamoxifen \sqsubseteq \exists metabolizedTo.(Compounds \sqcap \exists bindto.OestrogenReceptor),
   \\
   & & (\exists hasGene.CYP2D6) \sqcap (\exists TreatBy.Tamoxifen) \sqsubseteq \bot.
\end{eqnarray*}

These GCIs state that both tamoxifen and anti-aromatases are anti-estrogens;
tamoxifen can be metabolized into compounds which will bind to the oestrogen receptor;
and tamoxifen is contraindicated for people with the gene CYP2D6.

Suppose $Mary$ is a patient with the gene CYP2D6 and with some symptom captured by a concept $Symp$.
Then we can describe these information by an ABox $\mathcal{N}$ = $\{ Symp(Mary)$, $\exists hasGene.CYP2D6(Mary) \}$.
\qed
\end{example}

The semantics is defined by means of an \emph{interpretation}
$\mathcal{I}=(\Delta^{\mathcal{I}}, \cdot^{\mathcal{I}})$,
where the \emph{interpretation domain} $\Delta^{\mathcal{I}}$ is a non-empty set
composed of individuals, and $\cdot^{\mathcal{I}}$ is a
function which maps each concept name $A \in N_{C}$ to a set
$A^{\mathcal{I}}$ $\subseteq$ $\Delta^{\mathcal{I}}$,
maps each role name $r \in N_R$
to a binary relation $r^{\mathcal{I}}$ $\subseteq$
$\Delta^{\mathcal{I}}\times\Delta^{\mathcal{I}}$,
and maps each individual name $a \in
N_I$ to an individual $a^{\mathcal{I}}$ $\in$ $\Delta^{\mathcal{I}}$.
The function $\cdot^{\mathcal{I}}$ is inductively extended to arbitrary concepts as follows:
\begin{itemize}
  \item $\top := \Delta^{\mathcal{I}}$,
  \item $\bot := \emptyset$,
  \item $(C\sqcap D)^{\mathcal{I}} := C^{\mathcal{I}} \cap D^{\mathcal{I}}$, and
  \item $(\exists r.C)^{\mathcal{I}}$ $:=$
        $\{ x\in\Delta^{\mathcal{I}}$ $|$ there exists $y \in \Delta^{\mathcal{I}}$ such that
        $(x,y)\in r^{\mathcal{I}}$ and $y \in C^{\mathcal{I}} \}$.
\end{itemize}



The \emph{satisfaction relation} ``$\models$'' between any interpretation $\mathcal{I}$
and any GCI $C \sqsubseteq D$, concept assertion $C(a)$, role assertion $r(a,b)$,
TBox $\mathcal{T}$ or ABox $\mathcal{A}$ is defined inductively as follows:
$\mathcal{I}$ $\models$ $C \sqsubseteq D$ iff $C^{\mathcal{I}} \subseteq D^{\mathcal{I}}$;
$\mathcal{I}$ $\models$ $C(a)$ iff $a^{\mathcal{I}}$ $\in$ $C^{\mathcal{I}}$;
$\mathcal{I}$ $\models$ $r(a,b)$ iff $(a^{\mathcal{I}}, b^{\mathcal{I}})$ $\in$ $r^{\mathcal{I}}$;
$\mathcal{I}$ $\models$ $\mathcal{T}$ iff $\mathcal{I}$ $\models$ $X$ for every $X$ $\in$ $\mathcal{T}$; and
$\mathcal{I}$ $\models$ $\mathcal{A}$ iff $\mathcal{I}$ $\models$ $X$ for every $X$ $\in$ $\mathcal{A}$.

$\mathcal{I}$ is a \emph{model} of a KB $\mathcal{K} = \langle \mathcal{T},\mathcal{A} \rangle$
if $\mathcal{I}$ $\models$ $\mathcal{T}$ and $\mathcal{I}$ $\models$ $\mathcal{A}$.
We use $mod(\mathcal{K})$ to denote the set of models of KB $\mathcal{K}$.
Two KBs $\mathcal{K}_{1}$ and $\mathcal{K}_{2}$ are \emph{equivalent} (written $\mathcal{K}_{1} \equiv \mathcal{K}_{2}$)
iff $mod(\mathcal{K}_{1})$ = $mod(\mathcal{K}_{2})$.



There are many inference problems on DLs. Here we only introduce \emph{consistency} and \emph{entailment}.
A KB $\mathcal{K}$ = $\langle \mathcal{T},\mathcal{A} \rangle$ is \emph{consistent}
(or $\mathcal{A}$ is \emph{consistent} w.r.t. $\mathcal{T}$)
if $mod(\mathcal{K}) \neq \emptyset$.
%
$\mathcal{K}$ \emph{entails} a GCI, assertion or ABox $X$
(written $\mathcal{K}$ $\models$ $X$)
if $\mathcal{I}$ $\models$ $X$ for every $\mathcal{I} \in mod(\mathcal{K})$.
%
%
%
For these two inference problems the following results hold:
a KB $\mathcal{K}$ is inconsistent iff $\mathcal{K}$ $\models$ $\top \sqsubseteq \bot$
iff $\mathcal{K}$ $\models$ $\bot(a)$ for some individual name $a$ occurring in $\mathcal{K}$.
If $\mathcal{K}$ is inconsistent, then we say that $\mathcal{K}$ \emph{entails a clash}.

\begin{example} \label{example:inconsistence}
Consider the TBox $\mathcal{T}$ and ABox $\mathcal{N}$ presented in Example \ref{example:cancer treatment}.
Suppose there is an ABox $\mathcal{A}$ = $\{ TreatBy(Mary, y), Tamoxifen(y) \}$
and a KB $\mathcal{K} = \langle \mathcal{T}, \mathcal{A} \cup \mathcal{N} \rangle$.
It is obvious that $\mathcal{K}$ is inconsistent and entails a clash. More precisely, we have that
$\mathcal{K}$ $\models$ $\top \sqsubseteq \bot$ and
$\mathcal{K}$ $\models$ $\bot(Mary)$.
\qed
\end{example}



Let $X$ be a concept, GCI, assertion, TBox, ABox or KB.
Then $N_{C}^{X}$ (resp., $N_{R}^{X}$, $N_{I}^{X}$)
is the set of concept names (resp., role names, individual names)
occurring in $X$, and $sig(X)$ =
$N_{C}^{X} \cup N_{R}^{X} \cup N_{I}^{X}$.

For any concept $C$, the \emph{role depth} $rd(C)$ is the maximal nesting depth of $\exists$ in $C$.
Let $X$ be a GCI, assertion, TBox, ABox or KB.
Then $sub(X)$ is the set of all subconcepts occurring in $X$,
and $depth(X) = max \{ rd(C) \; | \; C \in sub(X) \}$.

\section{Formalization of Adaptation Based on $\mathcal{EL_{\bot}}$}  \label{section:ABox evolution of EL}


In this section we present a formalism for adaptation based on $\mathcal{EL_{\bot}}$.
There are many different approaches for the formalization of adaptation in CBR.
Here we follow the approach presented in \cite{lieber:07} to formulate adaptation as knowledge base revision,
with the difference that our formalism is based on the DL $\mathcal{EL_{\bot}}$
instead of propositional logic.

The basic idea of CBR is to solve similar problems with similar solutions.
The new problem that needs to be solved is called \emph{target problem}.
The problems which have been solved and stored are called \emph{source problems}.
Each source problem $srce$ has a \emph{solution} $sol_{srce}$,
and the pair $(srce, sol_{srce})$ is called a \emph{source case}.
A finite set of source cases forms a \emph{case base}.
%
Given a target problem $tgt$, the retrieval step of CBR will pick out a
source case $(srce, sol_{srce})$ according to the similarity between target problem and source problem,
and the adaptation step will generate a solution $sol_{tgt}$ for $tgt$ by adapting $sol_{srce}$.

In \cite{lieber:07}, the adaptation process is modeled as KB revision in propositional logic.
More precisely, let two formulas $kb_{1} = srce \wedge sol_{srce}$ and $kb_{2} = tgt$, then the
solution $sol_{tgt}$ is generated by calculating $(dk \wedge kb_{1}) \circ (dk \wedge kb_{2})$,
where $dk$ is the domain knowledge, and $\circ$ is a revision operator that satisfies the AGM postulates in propositional logic.


Here, after introducing the DL $\mathcal{EL_{\bot}}$ into CBR,
knowledge in a CBR system is composed of three parts:
\begin{itemize}
  \item the \emph{domain or background knowledge} which is represented as a TBox $\mathcal{T}$;

  \item the knowledge about \emph{case base} in which each \emph{source case} is described by a pair of ABoxes
$(srce, sol)$, where $srce$ describes the \emph{source problem}
and $sol$ describes the \emph{solution};

  \item the knowledge about \emph{target problem} described by an ABox $tgt$.

\end{itemize}

With such a framework, given a target problem $tgt$,
we can make use of similarity-measuring algorithms presented in the literature \cite{sanchez:11,lehmann:12} to
select a source case $(srce, sol)$ such that, by treating individual names occurring in $srce$
as variables, there exists a substitution $\sigma$ such that $\sigma(srce)$ and $tgt$ has the maximum similarity.
The retrieval algorithm applies $\sigma$ on $sol$ and return $\sigma(sol)$ as a \emph{possible solution} for the target problem.
Since retrieval algorithm is not the topic of this paper, we do not discuss it in detail here.

Now suppose a possible solution has been returned by the retrieval algorithm,
we defined adaptation setting as follows.

%



%

\begin{definition} \label{def:adaptation setting}
An \emph{adaptation setting} based on $\mathcal {EL_{\bot}}$ is a triple
$\mathcal{AS} = (\mathcal{T},\mathcal{A},\mathcal{N})$,
where $\mathcal{T}$ is a TBox describing the domain knowledge of the CBR system,
$\mathcal{N}$ is an ABox describing the target problem, and
$\mathcal{A}$ is an ABox describing the possible solution returned by retrieval algorithm.
%

An ABox $\mathcal{A}'$ is a \emph{solution} for an adaptation setting
$\mathcal{AS} = (\mathcal{T},\mathcal{A},\mathcal{N})$ if it satisfies the following requirements:
\begin{description}
  \item[(R1)] $\langle \mathcal{T},\mathcal{A}' \rangle$ $\models$ $\mathcal{N}$;

  \item[(R2)] $\mathcal{A}'$ $=$ $\mathcal{A} \cup \mathcal{N}$ if $\mathcal{A} \cup \mathcal{N}$ is consistent w.r.t. $\mathcal{T}$;

  \item[(R3)] if $\mathcal{N}$ is consistent w.r.t. $\mathcal{T}$
              then $\mathcal{A}'$ is also consistent w.r.t. $\mathcal{T}$.
%
%
\end{description}
\end{definition}

The adaptation setting defined above is similar to the \emph{instance-level revision setting} based on DLs \cite{calvanese:10};
\textbf{R1}-\textbf{R3} are just the basic requirements specified by the AGM postulates on revision operators \cite{alchourron:85,katsuno:91}.
More precisely,
\textbf{R1} specifies that a revision result must entail the new information $\mathcal{N}$;
\textbf{R2} states that the revision operator should not change the KB
$\langle \mathcal{T},\mathcal{A} \cup \mathcal{N} \rangle$ if there is no conflict;
\textbf{R3} states that the revision operator must preserve the consistency of KBs.

From the point of view of adaptation,
the requirements on solutions are explained as follows \cite{lieber:07}.
If \textbf{R1} is violated,
then it means that the adaptation process failed to solve the target problem.
\textbf{R2} states that if the possible solution
does not contradict the target problem w.r.t. the background knowledge,
then it can be applied directly to the target problem.
\textbf{R3} states that whenever the description of the target problem is consistent w.r.t. the domain knowledge,
the adaptation process provides satisfiable result.

In the literature, there exist some revision operators and algorithms that can generate revision results satisfying the above requirements
\cite{calvanese:10,flouris:05,kharlamov:13,lenzerini:11,wiener:06}.
However, in practice, besides these three necessary requirements,
we hope that the adaptation algorithm satisfies two more requirements.

Firstly, the adaptation algorithm should be syntax-independent.
More precisely,
if the descriptions of two target problems are essentially the same w.r.t. the domain knowledge,
then we hope that the adaptation process will generate the same solution for these two problems.
Under the framework of adaptation setting, we formalize this requirement as follows:
\begin{description}
  \item[(R4)] for any adaptation setting $\mathcal{AS}_{1} = (\mathcal{T},\mathcal{A}_{1},\mathcal{N}_{1})$
              satisfying $\langle \mathcal{T}, \mathcal{A}_{1} \rangle  \equiv  \langle \mathcal{T}, \mathcal{A} \rangle$
              and $\langle \mathcal{T}, \mathcal{N}_{1} \rangle  \equiv  \langle \mathcal{T}, \mathcal{N} \rangle$,
              there always exists a solution $\mathcal{A}'_{1}$ such that
              $\langle \mathcal{T}, \mathcal{A}'_{1} \rangle  \equiv  \langle \mathcal{T}, \mathcal{A}' \rangle$.
\end{description}

Secondly, the adaptation algorithm should guarantee a minimal change
so that the experience contained in the solution of source cases is preserved as much as possible.
Taking the adaptation setting $\mathcal{AS} = (\mathcal{T},\mathcal{A},\mathcal{N})$ as an example,
if we do not introduce the requirement on minimal change,
then in the case that $\mathcal{A} \cup \mathcal{N}$ is inconsistent w.r.t. $\mathcal{T}$,
the ABox $\mathcal{N}$ is a solution according to the definition.
However, it is not a “good” solution obviously, since information contained in $\mathcal{A}$ is completely lost.

We hope to specify the requirement on minimal change formally. However,
it is non-trivial to do it in a framework based on DLs.
One reason is that there are different approaches to define minimality,
and it is well-accepted that there is no general notion of minimality that will “do the
right thing” under all circumstances \cite{calvanese:10}.
Therefore, under the framework of adaptation setting, we only specify this requirement as follows:
\begin{description}
  \item[(R5)] the change from the KB $\langle \mathcal{T},\mathcal{A} \rangle$
              to the KB $\langle \mathcal{T},\mathcal{A}' \rangle$ is minimal.
\end{description}

To sum up, given an adaptation setting, we hope to generate a solution which not only satisfies
\textbf{R1}-\textbf{R3} specified in Definition \ref{def:adaptation setting},
but also satisfies \textbf{R4} and some reading of \textbf{R5}.



Before the end of this section, we look an example of adaptation setting.

\begin{example} \label{example:cancer treatment AS}
Consider the TBox $\mathcal{T}$ and ABox $\mathcal{N}$ presented in Example \ref{example:cancer treatment}.
Suppose $\mathcal{N}$ is a description of the target problem.
Suppose many successful treatment cases are recorded in the case base,
and from them a possible solution
$\mathcal{A}$ = $\{ TreatBy(Mary, y), Tamoxifen(y) \}$
is returned by the retrieval algorithm.
Then we get an adaptation setting $\mathcal{AS} = (\mathcal{T},\mathcal{A},\mathcal{N})$.
\qed
\end{example}

\section{Existing Approaches to Instance-Level Revision}  \label{section:Different Approaches}

As we mentioned in Section \ref{section:introduction}, there exist two groups of revision operators and algorithms
for DLs in the literature.
In this section, we show that they either do not support the DL $\mathcal {EL_{\bot}}$
or do not satisfy \textbf{R4} and \textbf{R5}.


\subsection{Model-based Approaches}

MBAs define revision operators over the distance between interpretations \cite{kharlamov:13}.
In propositional logic, it is easy to measure the distance between interpretations and
to calculate the revision results based on the distance \cite{katsuno:91}, since
each interpretation is only a truth assignment on propositional symbols.
However, in DLs,
it becomes very complex.



For an adaptation setting $\mathcal{AS} = (\mathcal{T},\mathcal{A},\mathcal{N})$,
let $\mathcal{M}$ be the set of all interpretations that satisfy both the solutions and the TBox $\mathcal{T}$.
Then, with MBAs, $\mathcal{M}$ is the set of models of
$\langle \mathcal{T},\mathcal{N} \rangle$
that are minimally distant from the models of $\langle \mathcal{T},\mathcal{A} \rangle$ \cite{kharlamov:13},
i.e.,
\begin{eqnarray*}
    \mathcal{M} = \{ \mathcal{J} \in mod(\langle \mathcal{T},\mathcal{N} \rangle)  \; | \;
                                      there \; exists \;  \mathcal{I} \in mod(\langle \mathcal{T},\mathcal{A} \rangle)
                                      \; such \; that \;
                                     dist(\mathcal{I},\mathcal{J}) =
                                    \\   min\{ dist(\mathcal{I}',\mathcal{J}') \; | \;
                                      \mathcal{I}' \in mod(\langle \mathcal{T},\mathcal{A} \rangle), \mathcal{J}' \in mod(\langle \mathcal{T},\mathcal{N} \rangle) \} \: \}.
\end{eqnarray*}

Let $\Sigma$ be the set of concept names and role names occurring in $\mathcal{AS}$.
There are four different approaches for measuring the distance $dist(\mathcal{I},\mathcal{J})$:
\begin{itemize}
  \item $dist^{s}_{\sharp}(\mathcal{I},\mathcal{J})$ = $\sharp \{ X \in \Sigma$ $|$
        $ X^{\mathcal{I}} \neq X^{\mathcal{J}} \}$,
  \item $dist^{s}_{\subseteq}(\mathcal{I},\mathcal{J})$ =
        $\{ X \in \Sigma $ $|$
        $ X^{\mathcal{I}} \neq X^{\mathcal{J}} \}$,
  \item $dist^{a}_{\sharp}(\mathcal{I},\mathcal{J})$ =
        $\mathop{sum} \limits_{X \in \Sigma}$
        $\sharp (X^{\mathcal{I}} \ominus X^{\mathcal{J}})$,
  \item $dist^{a}_{\subseteq}(\mathcal{I},\mathcal{J},X)$ =
        $X^{\mathcal{I}} \ominus X^{\mathcal{J}}$ for every $X \in \Sigma$,
\end{itemize}
where $X^{\mathcal{I}} \ominus X^{\mathcal{J}}$ =
$(X^{\mathcal{I}} - X^{\mathcal{J}}) \cup (X^{\mathcal{J}} - X^{\mathcal{I}})$.
Distances under $dist^{s}_{\sharp}$ and $dist^{a}_{\sharp}$
are natural numbers and are compared in the standard way.
Distances under $dist^{s}_{\subseteq}$ are sets and are compared by
set inclusion.
Distances under $dist^{a}_{\subseteq}$ are compared as follows:
$dist^{a}_{\subseteq}(\mathcal{I}_{1},\mathcal{J}_{1})$ $\leq$
$dist^{a}_{\subseteq}(\mathcal{I}_{2},\mathcal{J}_{2})$
iff
$dist^{a}_{\subseteq}(\mathcal{I}_{1},\mathcal{J}_{1},X)$ $\subseteq$
$dist^{a}_{\subseteq}(\mathcal{I}_{2},\mathcal{J}_{2},X)$ for every $X \in \Sigma$.
It is assumed that
the interpretations of individual names are fixed.
In \cite{kharlamov:13}, the above four different semantics for MBAs are denoted
as $\mathcal{G}^{s}_{\sharp}$, $\mathcal{G}^{s}_{\subseteq}$,
$\mathcal{G}^{a}_{\sharp}$, and $\mathcal{G}^{a}_{\subseteq}$ respectively.

Under the framework of adaptation setting,
we need to find a finite number of ABoxes $\mathcal{A}'_{i}$ $(1 \leq i \leq n)$ such that
$sig(\mathcal{A}'_{i})$ $\subseteq$ $sig(\mathcal{T})$ $\cup$ $sig(\mathcal{A})$ $\cup$ $sig(\mathcal{N})$
and
$\mathcal{M}$ =
$\mathop{\bigcup} \limits_{1 \leq i \leq n}$
        $mod(\langle \mathcal{T},\mathcal{A}'_{i} \rangle)$.

%

\begin{example} \label{example:model-based}
Consider an adaptation setting $\mathcal{AS}_{1}$ = $(\mathcal{T}_{1},\mathcal{A}_{1},\mathcal{N}_{1})$, where
\begin{eqnarray*}
    \mathcal{T}_{1} = \{ A \sqsubseteq \exists R.A, \;
                         A \sqsubseteq C, \;
                         E \sqcap \exists R.A \sqsubseteq \bot \}, \;\;
    \mathcal{A}_{1} = \{ A(a) \}, \;\;
    \mathcal{N}_{1} = \{ E(a) \}.
\end{eqnarray*}

Let $\Sigma = \{ A, C, E, R \}$.
Firstly, we investigate the semantics $\mathcal{G}^{s}_{\subseteq}$ and $\mathcal{G}^{s}_{\sharp}$.
It is obvious that, for any interpretations $\mathcal{I} \in mod(\langle \mathcal{T}_{1},\mathcal{A}_{1} \rangle)$ and
$\mathcal{J} \in mod(\langle \mathcal{T}_{1},\mathcal{N}_{1} \rangle)$,
it must be $A^{\mathcal{I}} \neq A^{\mathcal{J}}$
and $E^{\mathcal{I}} \neq E^{\mathcal{J}}$. Therefore, $\{ A, E \}$ is the minimal set of signatures
whose interpretations must be changed. So, under both $\mathcal{G}^{s}_{\subseteq}$ and $\mathcal{G}^{s}_{\sharp}$, we have that
\begin{eqnarray*}
    \mathcal{M} =
    \{ \mathcal{J} \in mod(\langle \mathcal{T}_{1},\mathcal{N}_{1} \rangle) \;
      | & &  \; there \; exists \; \mathcal{I} \in mod(\langle \mathcal{T}_{1},\mathcal{A}_{1} \rangle) \; such \; that
\\    & &  X^{\mathcal{I}} = X^{\mathcal{J}} \; for \; any \; X \in \Sigma \setminus \{ A, E \}
    \}.
\end{eqnarray*}

Now, for every positive integer $k$, let us construct an interpretation
$\mathcal{I}_{k}=(\Delta^{\mathcal{I}_{k}}, \cdot^{\mathcal{I}_{k}})$ as
$\Delta^{\mathcal{I}_{k}}$ = $\{ p_{1}, ..., p_{k} \}$,
$a^{\mathcal{I}_{k}}$ = $p_{1}$,
$A^{\mathcal{I}_{k}} = \emptyset$,
$E^{\mathcal{I}_{k}}$ = $\{ p_{1} \}$,
$C^{\mathcal{I}_{k}}$ = $\{ p_{1}, ..., p_{k} \}$, and
$R^{\mathcal{I}_{k}}$ = $\{ (p_{1},p_{2})$, ..., $(p_{k-1},p_{k})$, $(p_{k},p_{k}) \}$.
At the same time, be corresponding to each $\mathcal{I}_{k}$,
construct another interpretation $\mathcal{I}_{k}'=(\Delta^{\mathcal{I}_{k}'}, \cdot^{\mathcal{I}_{k}'})$ as
$\Delta^{\mathcal{I}_{k}'}$ = $\Delta^{\mathcal{I}_{k}}$,
$a^{\mathcal{I}_{k}'}$ = $p_{1}$,
$A^{\mathcal{I}_{k}'}$ = $\{ p_{1}, ..., p_{k} \}$,
$E^{\mathcal{I}_{k}'}$ = $\emptyset$,
$C^{\mathcal{I}_{k}'}$ = $C^{\mathcal{I}_{k}}$, and
$R^{\mathcal{I}_{k}'}$ = $R^{\mathcal{I}_{k}}$.
Then, it is obvious that $\mathcal{I}_{k} \in mod(\langle \mathcal{T}_{1},\mathcal{N}_{1} \rangle)$,
$\mathcal{I}_{k}' \in mod(\langle \mathcal{T}_{1},\mathcal{A}_{1} \rangle)$,
and $\mathcal{I}_{k} \in \mathcal{M}$.

In the case that $k = 1$, we can construct an ABox $\mathcal{A}'_{1}$ = $\{ E(a)$, $C(a)$, $R(a,a) \}$
so that $\mathcal{I}_{1}$ is contained in
$mod(\langle \mathcal{T}_{1},\mathcal{A}'_{1} \rangle)$.
However, for any $k \geq 2$, it is impossible to find an ABox $\mathcal{A}'_{i}$ such that
$sig(\mathcal{A}'_{i})$ $\subseteq$ $\Sigma \cup \{ a \}$ and
$\mathcal{I}_{k}$ $\in$
$mod(\langle \mathcal{T}_{1},\mathcal{A}'_{i} \rangle)$.
So, solutions of adaptation under $\mathcal{G}^{s}_{\subseteq}$ and $\mathcal{G}^{s}_{\sharp}$ is not expressible.


Secondly, we investigate the semantics $\mathcal{G}^{a}_{\subseteq}$ and $\mathcal{G}^{a}_{\sharp}$.
It is obvious that, for any interpretations $\mathcal{I} \in mod(\langle \mathcal{T}_{1},\mathcal{A}_{1} \rangle)$ and
$\mathcal{J} \in mod(\langle \mathcal{T}_{1},\mathcal{N}_{1} \rangle)$,
it must be $a^{\mathcal{I}} \notin E^{\mathcal{I}}$
and $a^{\mathcal{J}} \notin A^{\mathcal{J}}$. Therefore $\{ A(a), E(a) \}$ is the minimal set of atoms
whose interpretations must be changed. So, under both $\mathcal{G}^{a}_{\subseteq}$ and $\mathcal{G}^{a}_{\sharp}$, we have that
\begin{eqnarray*}
    \mathcal{M} =
    \{ \mathcal{J} \in mod(\langle \mathcal{T}_{1},\mathcal{N}_{1} \rangle) \;
      & & | \; there \; exists \; \mathcal{I} \in mod(\langle \mathcal{T}_{1},\mathcal{A}_{1} \rangle) \; such \; that \;
\\    & &   A^{\mathcal{I}} \ominus A^{\mathcal{J}} =
            E^{\mathcal{I}} \ominus E^{\mathcal{J}} = \{ a^{\mathcal{I}} \}, \;
            and
\\    & &   X^{\mathcal{I}} = X^{\mathcal{J}} \; for \; any \; X \in \Sigma \setminus \{A, E\}
    \}.
\end{eqnarray*}

From $\mathcal{M}$ we can construct an ABox
$\mathcal{A}_{1}'$ = $\{ E(a), C(a), R(a,a) \}$
so that $\mathcal{M}$ = $mod(\langle \mathcal{T}_{1}, \mathcal{A}_{1}' \rangle)$.
Therefore, under both $\mathcal{G}^{a}_{\subseteq}$ and $\mathcal{G}^{a}_{\sharp}$,
$\mathcal{A}_{1}'$ is a solution for the adaptation setting.
This result is very strange,
since during the adaptation process there seems to be no ``good'' reason to enforce the assertion $R(a,a)$ to hold.
\qed
\end{example}

To sum up, there are four notions of computing models in existing MBAs.
For the adaptation based on $\mathcal{EL_{\bot}}$,
two notions suffer from inexpressibility and the other two notions are semantically questionable.

\subsection{Formula-based Approaches}

In the literature there are two typical formula-based approaches for instance-level revision in DLs.

The first one is based on deductive closures \cite{calvanese:10,lenzerini:11}.
Under the framework studied in our paper,
given an adaptation setting $\mathcal{AS} = (\mathcal{T},\mathcal{A},\mathcal{N})$,
this approach will firstly calculates the deductive closure of $\mathcal{A}$ w.r.t. $\mathcal{T}$
(denoted $cl_{\mathcal{T}}(\mathcal{A})$);
then it computes a maximal subset $\mathcal{A}_{m}$ of $cl_{\mathcal{T}}(\mathcal{A})$ that
does not conflict with $\mathcal{N}$ and $\mathcal{T}$; and finally returns $\mathcal{A}_{m}$ $\cup$ $\mathcal{N}$ as a solution.
Such an approach behaves well in DLs of the \emph{DL-Lite} family,
where $cl_{\mathcal{T}}(\mathcal{A})$ is finite and can be calculated effectively.
However, it does not work in $\mathcal{EL_{\bot}}$,
since $cl_{\mathcal{T}}(\mathcal{A})$ is infinite in $\mathcal{EL_{\bot}}$
and can not be calculated directly.
Therefore, it can not be applied for solving our problem.

The second FBA is based on justifications (also known as MinAs or kernel) \cite{wiener:06}.
Under the framework studied here,
given an adaptation setting $\mathcal{AS} = (\mathcal{T},\mathcal{A},\mathcal{N})$,
this approach will firstly construct a KB $\mathcal{K}_{0} = \langle \mathcal{T},\mathcal{A} \cup \mathcal{N} \rangle$,
and find all the minimal subsets of $\mathcal{K}_{0}$ that entail a clash (i.e., all justifications for clashes);
then it will compute a minimal set $\mathcal{R} \subseteq \mathcal{A}$ which contains at least one element from each justification
(such a set is also called a \emph{repair});
and finally returns $(\mathcal{A} \cup \mathcal{N}) \setminus \mathcal{R}$ as a solution.
This approach is applicable to DLs such as $\mathcal{SHOIN}$,
and obviously can deal with $\mathcal{EL_{\bot}}$.
However, as shown by the following examples, it is syntax-dependent
and not fine-grained, and therefore does not satisfy our requirements specified by \textbf{R4} and \textbf{R5}.

\begin{example}  \label{example:formula-based-1}
Consider the adaptation setting $\mathcal{AS}_{1}$ = $(\mathcal{T}_{1},\mathcal{A}_{1},\mathcal{N}_{1})$
described in the previous example.
It is obvious that $\langle \mathcal{T}_{1}, \mathcal{A}_{1} \cup \mathcal{N}_{1} \rangle$
$\models \bot(a)$ and for which there is only one justification
$\mathcal{J}$ = $\{ A \sqsubseteq \exists R.A, E \sqcap \exists R.A \sqsubseteq \bot, A(a), E(a) \}$.

Since $\mathcal{T}_{1}$ is fixed and $E(a) \in \mathcal{N}_{1}$,
the only choice is to remove $A(a)$ from $\mathcal{A}_{1} \cup \mathcal{N}_{1}$
and get the solution $\mathcal{A}_{1}' = \{ E(a) \}$.
This result is not so good, since it loses many information which is entailed by
the KB $\langle \mathcal{T}_{1}, \mathcal{A}_{1} \rangle$
and not conflicted with $\mathcal{N}_{1}$,
such as the concept assertions $C(a)$ and $\exists R.C(a)$.
In other words, this result does not satisfy \textbf{R5}.
\qed
\end{example}

\begin{example}  \label{example:formula-based-2}
Consider another adaptation setting $\mathcal{AS}_{2}$ = $(\mathcal{T}_{2},\mathcal{A}_{2},\mathcal{N}_{2})$,
where $\mathcal{T}_{2}$ = $\mathcal{T}_{1}$, $\mathcal{N}_{2}$ = $\mathcal{N}_{1}$
and $\mathcal{A}_{2} = \{ A(a), C(a), \exists R.C(a) \}$.
It is obvious that
$\langle \mathcal{T}_{1}, \mathcal{A}_{1} \rangle$ $\equiv$ $\langle \mathcal{T}_{2}, \mathcal{A}_{2} \rangle$
and $\langle \mathcal{T}_{1}, \mathcal{N}_{1} \rangle$ $\equiv$ $\langle \mathcal{T}_{2}, \mathcal{N}_{2} \rangle$.


Now, apply the FBA based on justifications again,
we will get a solution
$\mathcal{A}_{2}' = \{ E(a), C(a), \exists R.C(a) \}$.
This solution is essentially different from the solution $\mathcal{A}_{1}' = \{ E(a) \}$ of $\mathcal{AS}_{1}$,
since $\langle \mathcal{T}_{1}, \mathcal{A}_{1}' \rangle$ $\not\equiv$ $\langle \mathcal{T}_{2}, \mathcal{A}_{2}' \rangle$.
So, the FBA based on justifications does not satisfy \textbf{R4}.
It is unhelpful that we get two different solutions for two target problems which are essentially the same.
%
\qed
\end{example}

To sum up, for the adaptation based on $\mathcal{EL_{\bot}}$,
existing FBAs either can not be applied directly, or can be applied but is syntax-dependent and not fine-grained.


\section{Our Approach for Adaptation Based on $\mathcal {EL_{\bot}}$}  \label{section:revision algorithm}

In this section we present an algorithm for adaptation based on $\mathcal {EL_{\bot}}$.
Our algorithm is based on a structure named revision graph,
which is close to the completion graph used in classical tableau decision algorithms of DLs \cite{horrocks:07}.
We firstly introduce some notions and operations on this structure,
then present the algorithm, and finally discuss two examples.




\subsection{Notions and Operations on Revision Graph} \label{subsection:revision graph}


A \emph{revision graph} for $\mathcal {EL_{\bot}}$ is a directed graph $\mathcal{G}$ = $(V,E,\mathcal{L})$, where
\begin{itemize}
  \item $V$ is a finite set of nodes composed of individual names and variables;
  \item $E \subseteq V \times V$ is a set of edges satisfying:
     \begin{itemize}
       \item there is no edge from variables to individual names, and
       \item for each variable $y$ $\in$ $V$, there is at most one node
             $x$ with $\langle x,y \rangle $ $\in$ $E$;
     \end{itemize}
  \item each node $x$ $\in$ $V$ is labelled with a set of concepts $\mathcal{L}(x)$; and
  \item each edge $\langle x,y \rangle $ $\in$ $E$ is labelled with a set of role names $\mathcal{L}(\langle x,y \rangle )$;
        furthermore, if $y$ is a variable then $\sharp \mathcal{L}(\langle x,y \rangle ) = 1$.
\end{itemize}

For each edge $\langle x,y \rangle $ $\in$ $E$, we call $y$ a \emph{successor} of $x$ and $x$ a \emph{predecessor} of $y$.
\emph{Descendant} is the transitive closure of successor.

For any node $x$ $\in$ $V$, we use $level(x)$ to denote the level of $x$ in the graph, and define it inductively as follows:
$level(x)$ $=$ 0 if $x$ is an individual name, and
$level(x)$ $=$ $level(y)+1$ if $x$ is a variable with a predecessor $y$.

A graph $\mathcal{B}=(V',E',\mathcal{L}')$ is a \emph{branch}
of $\mathcal{G}$ if $\mathcal{B}$ is a tree and a subgraph of $\mathcal{G}$.

A branch $\mathcal{B}_{1} = (V_{1},E_{1},\mathcal{L}_{1})$ is \emph{subsumed} by another branch $\mathcal{B}_{2} = (V_{2},E_{2},\mathcal{L}_{2})$
if $\mathcal{B}_{1}$ and $\mathcal{B}_{2}$ have the same root node, $\sharp(V_{1} \cap V_{2}) = 1$, and
there is a function $f: V_{1} \rightarrow V_{2}$ such that:
$f(x)$ $=$ $x$ if $x$ is the root node,
$\mathcal{L}_{1}(x)$ $\subseteq$ $\mathcal{L}_{2}(f(x))$ for every node $x \in V_{1}$,
$\langle f(x),f(y) \rangle$ $\in$ $E_{2}$ for every edge $\langle x,y \rangle$ $\in E_{1}$, and
$\mathcal{L}_{1}(\langle x,y \rangle)$ $\subseteq$ $\mathcal{L}_{2}(\langle f(x),f(y) \rangle)$
        for every edge $\langle x,y \rangle$ $\in E_{1}$.

A branch $\mathcal{B}$ is \emph{redundant} in $\mathcal{G}$ if
every node in $\mathcal{B}$ except the root is a variable, and
$\mathcal{B}$ is subsumed by another branch in $\mathcal{G}$.

Revision graphs can be seen as ABoxes with variables.
Given a revision graph $\mathcal{G}$ = $(V,E,\mathcal{L})$, we call 
    $\mathcal{A}_{\mathcal {G}} = \mathop {\bigcup}\limits_{x \in V} \{C(x) \; | \; C \in \mathcal{L}(x) \}
    \; \cup \mathop {\bigcup}\limits_{\langle x,y \rangle  \in E} \{R(x,y) \; | \; R \in \mathcal{L}(\langle x,y \rangle ) \}$
as the \emph{ABox representation} of $\mathcal{G}$,
and call $\mathcal{G}$ as the \emph{revision-graph representation} of
$\mathcal{A}_{\mathcal{G}}$.

Given a KB $\mathcal {K}=\langle \mathcal{T},\mathcal{A} \rangle$ and a non-negative integer $k$,
we use procedure B-MW($\mathcal {K}$, $k$) to construct a revision graph for them.


\vspace{-0.6cm}
\begin{procedure} 
\KwIn{ a KB $\mathcal {K}=\langle \mathcal{T},\mathcal{A} \rangle$ and a non-negative integer $k$.}
\KwOut{a revision graph $\mathcal{G}$ = $(V,E,\mathcal{L})$.}

\nl Initialize the revision graph $\mathcal{G}$ = $(V,E,\mathcal{L})$ as
\begin{itemize}
  \item $V = N_{I}^{\mathcal {K}}$,
  \item $\mathcal{L}(a) = \{ C  \; | \; C(a) \in \mathcal{A} \}$ for each node $a \in V$,
  \item $E = \{ \langle a,b \rangle   \; |
            \; there \; is \; some \; R \; with \; R(a,b) \in \mathcal{A} \}$,
  \item $\mathcal{L}(\langle a,b \rangle ) = \{ R  \; | \; R(a,b) \in \mathcal{A} \}$ for each edge $\langle a,b \rangle  \in E$.
\end{itemize}

\nl  \While{there exists an expansion rule in Fig. 1 that is applicable to $\mathcal{G}$}
  {
    expand $\mathcal{G}$ by applying this rule.
  }
%
%
%

\nl  \For{ each node $x \in V$ }
  {
   $\mathcal{L}(x)$ = $\{ C \in \mathcal{L}(x)$ $|$ $C$ is a concept name $\}$.
  }

\nl  \While{there exists a redundant branch $\mathcal{B} = (V_{\mathcal{B}},E_{\mathcal{B}},\mathcal{L}_{\mathcal{B}})$ in $\mathcal{G}$}
  {

   $E = E \setminus E_{\mathcal{B}}$;

   $V = V \setminus (V_{\mathcal{B}} \setminus \{ x_{\mathcal{B}} \})$, where $x_{\mathcal{B}}$ is the root of $\mathcal{B}$.
  }

\nl  Return $\mathcal{G}$ = $(V,E,\mathcal{L})$.

\caption{B-MW()($\mathcal {K}$, $k$)}
\end{procedure}
\vspace{-0.6cm}

It is easy to prove the following property.

%

\begin{proposition} \label{theorem:relationship between ABox and MNW}
Given a KB $\mathcal{K}$
and a non-negative integer $k$,
let $\mathcal{G}$ = B-MW($\mathcal {K}, k$),
and let $\mathcal{A}_{\mathcal{G}}$ be the ABox representation of $\mathcal{G}$.
Then, for any ABox assertion $\alpha$ with $sig(\alpha)$ $\subseteq$ $sig(\mathcal{K})$
and $depth(\alpha)$ $\leq$ $k$,
$\mathcal{K}$ $\models$ $\alpha$  iff  $\langle \emptyset,\mathcal{A}_{\mathcal{G}} \rangle$ $\models$ $\alpha$.
\end{proposition}
%

Based on this property, given an adaptation setting
$\mathcal{AS} = (\mathcal{T},\mathcal{A},\mathcal{N})$, we can firstly construct
a revision graph $\mathcal{G}$ = B-MW($\langle \mathcal{T},\mathcal{A} \rangle, k$)
for some well selected integer $k$,
and then realize adaptation by computing a maximal subset $\mathcal{A}_{m}$ of
$\mathcal{A}_\mathcal{G}$ that does not conflict with $\mathcal{N}$.

\begin{small}
\setlength{\unitlength}{0.9mm}
\begin{picture}(120,54)
\put(-6,5){\line(1,0){136}} 
\put(-6,5){\line(0,1){46}} 
\put(130,5){\line(0,1){46}} 
\put(-6,51){\line(1,0){136}} 
\put(-6,1){\textbf{Fig. 1.} Expansion rules used by the procedure B-MW($\mathcal {K}$, $k$).} 
\put(-5,47){GCI$_{I}$-rule:} %
\put(13,47){
   if $x$ $\in$ $N_{I}^{\mathcal{K}}$, $C \sqsubseteq D$ $\in$ $\mathcal{T}$,
   $D \notin \mathcal{L}(x)$, and
   $\langle \mathcal{T},\mathcal{A} \rangle$ $\models$ $C(x)$,
   } %
\put(14,42){then set $\mathcal{L}(x) = \mathcal{L}(x) \cup \{ D \}$.}
\put(-5,37){GCI$_{V}$-rule:} %
\put(13,37){
   if $x$ $\notin$ $N_{I}^{\mathcal {K}}$, $C \sqsubseteq D$ $\in$ $\mathcal{T}$,
   $D \notin \mathcal{L}(x)$, and
   $\langle \mathcal{T},\{ E(x) \; | \; E \in \mathcal{L}(x) \} \rangle$ $\models$ $C(x)$,
   } %
\put(14,32){then set $\mathcal{L}(x) = \mathcal{L}(x) \cup \{ D \}$.}
\put(1,27){$\sqcap$-rule:} %
\put(13,27){
   if $C_{1} \sqcap C_{2}$ $\in$ $\mathcal{L}(x)$, and $\{ C_{1}, C_{2} \} \nsubseteq \mathcal{L}(x)$,
   } %
\put(14,22){then set $\mathcal{L}(x) = \mathcal{L}(x) \cup \{ C_{1}, C_{2} \}$.}
\put(1,17){$\exists$-rule:} %
\put(13,17){
   if $\exists R.C$ $\in$ $\mathcal{L}(x)$, $x$ has no successor $z$ with $C \in \mathcal{L}(z)$,
   and $level(x)$ $\leq$ $k$,
   } %
\put(14,12){then introduce a new variable $z$, set $V = V \cup \{ z \}$, $E = E \cup \{ \langle x,z \rangle  \}$,}
\put(14,7){$\mathcal{L}(z) =$  $\{ C \}$, and $\mathcal{L}(\langle x,z \rangle ) = \{ R \}$.}
\end{picture}
\end{small}

Because $\mathcal{A}_{m}$ is an ABox that may contain variables, we use the following procedure
to roll up variables contained in it.

\vspace{-0.6cm}
\begin{procedure} [bht] 
\KwIn{ an ABox $\mathcal{A}_{m}$ that may contain variables, and a TBox $\mathcal{T}$. }
\KwOut{ an ABox $\mathcal{A}_{0}$ without variables.}

\nl Transform $\mathcal{A}_{m}$ into its revision-graph representation $\mathcal{G}$ = $(V,E,\mathcal{L})$.

\nl Delete from $V$ the variables which are not descendants of any individual name.

\nl  \While{there exists variable in $V$}
  {
    select a variable $y$ $\in$ $V$ that has no successor;

    $x$ = the predecessor of $y$;

    \textbf{if} $\mathcal{L}(y)$ $\neq$ $\emptyset$
    \textbf{then} $C_{y}$ = $\mathop {\bigsqcap}\limits_{C \in \mathcal{L}(y)} C$
    \textbf{else} $C_{y}$ = $\top$;


    $R$ = the role name contained in $\mathcal{L}(\langle x,y \rangle)$;

    \textbf{if} $\langle \mathcal{T},\{ D(x) \; | \; D \in \mathcal{L}(x) \} \rangle$ $\not\models$ $(\exists R.C_{y})(x)$
    \textbf{then} $\mathcal{L}(x)$ $=$ $\mathcal{L}(x)$ $\cup$ $\{ \exists R.C_{y} \}$;


    $E$ $=$ $E$ $\setminus$ $\{ \langle x,y \rangle  \}$;

    $V = V \setminus \{ y \}$.
  }

\nl  Return $\mathcal{A}_{0}$ = $\mathop {\bigcup}\limits_{x \in V} \{C(x) \; | \; C \in \mathcal{L}(x) \}
    \; \cup \mathop {\bigcup}\limits_{\langle x,y \rangle  \in E} \{R(x,y) \; | \; R \in \mathcal{L}(\langle x,y \rangle ) \}$.


\caption{Rolling()($\mathcal{A}_{m}$, $\mathcal{T}$)}
\end{procedure}
\vspace{-0.6cm}

\subsection{The Revision Algorithm} \label{subsection:revision}



Let $\mathcal{T}$ be a TBox, and let $\mathcal{A}$, $\mathcal{N}$ be two ABoxes.
If $\langle \mathcal{T}, \mathcal{A} \cup \mathcal{N} \rangle$ $\models$ $\top \sqsubseteq \bot$, then:
\begin{itemize}
  \item a set $\mathcal{J}$ $\subseteq$ $\mathcal{A}$ is a $(\mathcal{A},\mathcal{N})$\emph{-justification} for a clash w.r.t. $\mathcal{T}$
if
$\langle \mathcal{T}, \mathcal{J} \cup \mathcal{N} \rangle$
$\models$ $\top \sqsubseteq \bot$ and
$\langle \mathcal{T}, \mathcal{J}' \cup \mathcal{N} \rangle$ $\not\models$ $\top \sqsubseteq \bot$ for every $\mathcal{J}'$ $\subset$ $\mathcal{J}$;

  \item a set $\mathcal{R}$ $\subseteq$ $\mathcal{A}$ is a $(\mathcal{A},\mathcal{N})$\emph{-repair} for clashes w.r.t. $\mathcal{T}$
if
$\sharp(\mathcal{R} \cap \mathcal{J}) = 1$ for every $(\mathcal{A},\mathcal{N})$\emph{-justification} $\mathcal{J}$.
\end{itemize}

Now we are ready to present our algorithm for adaptation based on $\mathcal {EL_{\bot}}$.

\vspace{-0.6cm}
\begin{algorithm}[ht]
\KwIn{ an adaptation setting $\mathcal{AS} = (\mathcal{T},\mathcal{A},\mathcal{N})$, and a non-negative integer $k$.}
\KwOut{a finite number of pairs $(\mathcal{A}_{1}'', \mathcal{R}_{1})$, ..., $(\mathcal{A}_{n}'', \mathcal{R}_{n})$,
       where $\mathcal{A}_{i}''$ is a solution
      \newline
      and $\mathcal{R}_{i}$ records the information been removed.}

\eIf{$\mathcal{A}$ $\cup$ $\mathcal{N}$ is consistent w.r.t. $\mathcal{T}$}
{
return $(\mathcal{A} \cup \mathcal{N}, \emptyset)$;
}
{
$\mathcal{G}$ = B-MW($\langle \mathcal{T},\mathcal{A} \rangle, k$);

$\mathcal{A}_{\mathcal{G}}$ = the ABox representation of $\mathcal{G}$;


$S_\mathcal{R}$ = $\{ \mathcal{R}_{1}, ..., \mathcal{R}_{n} \}$
all the $(\mathcal{A}_{\mathcal{G}},\mathcal{N})$-repairs for a clash w.r.t. $\mathcal{T}$;

\For{$i \leftarrow 1$ \KwTo $n$}{

%
$\mathcal{A}_{i}$ = $\mathcal{A}_{\mathcal{G}} \setminus \mathcal{R}_{i}$;

%
$\mathcal{A}_{i}'$ = Rolling($\mathcal{A}_{i}$, $\mathcal{T}$);
}

return $(\mathcal{A}_{1}' \cup \mathcal{N}, \mathcal{R}_{1})$, ..., $(\mathcal{A}_{n}' \cup \mathcal{N}, \mathcal{R}_{n})$.
}

%
%
%
%
%
%
%
\caption{Adaptation($\mathcal{AS}$, $k$)}
\end{algorithm}
\vspace{-0.6cm}

Given an adaptation setting $\mathcal{AS} = (\mathcal{T},\mathcal{A},\mathcal{N})$,
our algorithm will firstly construct a revision graph $\mathcal{G}$ according to the
KB $\langle \mathcal{T},\mathcal{A} \rangle$ and some integer $k$, where $k$ is required to be greater than the
role depthes of all the concepts occurring in the adaptation setting.
Secondly, a revision process based on justifications will be carried out on the
ABox representation $\mathcal{A}_{\mathcal{G}}$ of $\mathcal{G}$.
Thirdly, for each maximal subset $\mathcal{A}_{i}$ of $\mathcal{A}_{\mathcal{G}}$ that
does not conflict with $\mathcal{N}$ and $\mathcal{T}$, the procedure
Rolling($\mathcal{A}_{i}$, $\mathcal{T}$) will be used to roll up variables and get an ABox
$\mathcal{A}_{i}'$.
Finally, the ABox $\mathcal{A}_{i}'$ $\cup$ $\mathcal{N}$ will be returned as a solution for the
adaptation setting; together with the solution, an ABox $\mathcal{R}_{i}$ which records the
information been removed from the solution of source problem is also displayed,
so that the user can determine the best solution according to it.

The following theorems state that our algorithm satisfies \textbf{R1}-\textbf{R4}.

\begin{theorem} \label{theorem:3 AGM postulates}
Let $(\mathcal{A}_{i}'', \mathcal{R}_{i})$ $(1 \leq i \leq n)$ be the pairs
returned by Adaptation$(\mathcal{AS}, k)$ for $\mathcal{AS} = (\mathcal{T},\mathcal{A},\mathcal{N})$.
Then the following statements hold for every $1 \leq i \leq n$:
\emph{(1)} $\langle \mathcal{T},\mathcal{A}_{i}'' \rangle$ $\models$ $\mathcal{N}$;
\emph{(2)} $\mathcal{A}_{i}''$ $=$ $\mathcal{A} \cup \mathcal{N}$
if $\mathcal{A} \cup \mathcal{N}$ is consistent w.r.t. $\mathcal{T}$; and
\emph{(3)} if $\mathcal{N}$ is consistent w.r.t. $\mathcal{T}$ then
$\mathcal{A}_{i}''$ is also consistent w.r.t. $\mathcal{T}$.
%
\end{theorem}


\begin{theorem} \label{theorem:syntax independent}
Given two adaptation settings $\mathcal{AS}_{i} = (\mathcal{T},\mathcal{A}_{i},\mathcal{N}_{i})$ $(i = 1, 2)$
and an integer $k$ satisfying $k \geq$ $depth(\mathcal{T})$, $k \geq$ $depth(\mathcal{A}_{i})$,
and $k \geq$ $depth(\mathcal{N}_{i})$ $(i = 1, 2)$.
Let $(\mathcal{A}_{1}'', \mathcal{R}_{1})$ be a pair returned by Adaptation$(\mathcal{AS}_{1}, k)$.
If $\langle \mathcal{T}, \mathcal{A}_{1} \rangle$ $\equiv$ $\langle \mathcal{T}, \mathcal{A}_{2} \rangle$ and
$\langle \mathcal{T}, \mathcal{N}_{1} \rangle$ $\equiv$ $\langle \mathcal{T}, \mathcal{N}_{2} \rangle$,
then there must be a pair $(\mathcal{A}_{2}'', \mathcal{R}_{2})$
returned by Adaptation$(\mathcal{AS}_{2}, k)$ such that
$\langle \mathcal{T}, \mathcal{A}_{1}'' \rangle$ $\equiv$ $\langle \mathcal{T}, \mathcal{A}_{2}'' \rangle$
and $\mathcal{R}_{2}$ = $\sigma(\mathcal{R}_{1})$ for some substitution $\sigma$ of variables.
%
\end{theorem}

Theorem \ref{theorem:syntax independent} is based on the following fact:
let $\mathcal{G}_{i}$ = B-MW($\langle \mathcal{T},\mathcal{A}_{i} \rangle,k$) $(i=1,2)$,
then $\mathcal{G}_{1}$ and $\mathcal{G}_{2}$ are identical up to variable renaming
in the case that $k$ is sufficiently large.
There is no requirement on the value of $k$ in Theorem \ref{theorem:3 AGM postulates}.

In our algorithm, the revision graph $\mathcal{G}$ constructed by the procedure B-MW($\langle \mathcal{T},\mathcal{A} \rangle, k$)
is in fact a non-redundant $k$-depth-bounded model for the KB $\langle \mathcal{T},\mathcal{A} \rangle$.
Therefore, our revision process works on fine-grained representation of models
and guarantees the minimal change principle in a fine-grained level.
So, our algorithm satisfies the property specified by \textbf{R5}.


The following theorem states that our algorithm is in exponential time.

\begin{theorem}
For any adaptation setting $\mathcal{AS} = (\mathcal{T},\mathcal{A},\mathcal{N})$,
assume the role depth of every concept occurring in
$\mathcal{AS}$ is bounded by some integer $k$,
then the algorithm Adaptation$(\mathcal{AS}, k)$
runs in time exponential with respect to the size of $\mathcal{AS}$.
\end{theorem}
%

\subsection{Examples}


\begin{example}
Consider the adaptation setting $\mathcal{AS}_{1}$ = $(\mathcal{T}_{1},\mathcal{A}_{1},\mathcal{N}_{1})$
described in Example \ref{example:model-based}.
Since $max\{ depth(\mathcal{T}_{1}), depth(\mathcal{A}_{1}), depth(\mathcal{N}_{1}) \}=1$,
we let $k$=1 and execute the algorithm Adaptation$(\mathcal{AS}_{1}, k)$.

Firstly we call the procedure B-MW($\langle \mathcal{T}_{1},\mathcal{A}_{1} \rangle, k$)
to construct a revision graph $\mathcal{G}$.
Two variables $x_{1}$, $x_{2}$ are introduced during this procedure.
By treating $\mathcal{G}$ as an ABox, we get
$\mathcal{A}_{\mathcal{G}} = \{  A(a), \: C(a), \: R(a,x_{1}), \: A(x_{1}),
                                  \: C(x_{1}), \: R(x_{1}, \: x_{2}), \: A(x_{2}), \: C(x_{2}) \}.
$

Secondly, for the clash $\mathcal{A}_{\mathcal{G}} \cup \mathcal{N}_{1}$ $\models_{\mathcal{T}_{1}}$ $\top \sqsubseteq \bot$,
there are two $(\mathcal{A}_{\mathcal{G}},\mathcal{N}_{1})$-justifications
$\mathcal{J}_{1}  =  \{ A \sqsubseteq \exists R.A,  \:  E \sqcap \exists R.A \sqsubseteq \bot,
                          \:  E(a), \: A(a) \}$ and
$\mathcal{J}_{2}  =  \{ E \sqcap \exists R.A \sqsubseteq \bot, \: E(a), \: A(a), \: R(a,x_{1}), \: A(x_{1}) \}$.
%
%
Based on them we get two $(\mathcal{A}_{\mathcal{G}},\mathcal{N}_{1})$-repairs
$\mathcal{R}_{1} = \{ A(a), A(x_{1}) \}$ and
$\mathcal{R}_{2} = \{ A(a), R(a,x_{1}) \}$.
%

Thirdly, from $\mathcal{R}_{1}$ we get
$\mathcal{A}_{1}$ = $\mathcal{A}_{\mathcal{G}} \setminus \mathcal{R}_{1}$ =
$\{  C(a)$, $R(a,x_{1})$, $C(x_{1})$, $R(x_{1}, x_{2})$, $A(x_{2})$, $C(x_{2}) \}$,
and then get $\mathcal{A}_{1}'$ = Rolling($\mathcal{A}_{1}$, $\mathcal{T}_{1}$)
= $\{ C(a)$, $\exists R.(C \sqcap \exists R.A)(a) \}$.
From $\mathcal{R}_{2}$, we get
$\mathcal{A}_{2}$ = $\mathcal{A}_{\mathcal{G}} \setminus \mathcal{R}_{2}$ =
$\{C(a)$, $A(x_{1})$, $C(x_{1})$, $R(x_{1},x_{2})$, $A(x_{2})$, $C(x_{2}) \}$
and $\mathcal{A}_{2}'$ = Rolling($\mathcal{A}_{2}$, $\mathcal{T}_{1}$)
= $\{ C(a) \}$.

Finally, the algorithm returns $(\mathcal{A}_{1}' \cup \mathcal{N}_{1}, \mathcal{R}_{1})$
and $(\mathcal{A}_{2}' \cup \mathcal{N}_{1}, \mathcal{R}_{2})$.

Now $\mathcal{R}_{2}$ contains a role assertion $R(a,x_{1})$ which indicates that
all the information related to $x_{1}$ is lost in $\mathcal{A}_{2}' \cup \mathcal{N}_{1}$,
hence $\mathcal{A}_{2}'$ is rather weak (in fact weaker than $\mathcal{A}_{1}'$).
Thus the user should select $(\mathcal{A}_{1}' \cup \mathcal{N}_{1}, \mathcal{R}_{1})$
and get a solution $\mathcal{A}_{1}''$ = $\mathcal{A}_{1}' \cup \mathcal{N}_{1}$
= $\{ E(a)$, $C(a), \: \exists R.(C \sqcap \exists R.A)(a) \}$.
%
%
\qed
\end{example}


Looking back Example \ref{example:model-based} and Example \ref{example:formula-based-1},
given the same adaptation setting $\mathcal{AS}_{1}$ = $(\mathcal{T}_{1},\mathcal{A}_{1},\mathcal{N}_{1})$,
the solution generated by MBAs under the semantics $\mathcal{G}^{a}_{\subseteq}$ and $\mathcal{G}^{a}_{\sharp}$
is $\mathcal{A}_{1}'  = \{ E(a), C(a), R(a,a) \}$;
the solution generated by the FBA based on justifications is $\mathcal{A}_{1}' = \{ E(a) \}$.
Obviously, our algorithm is more fine-grained.

\begin{example}
Consider the adaptation setting $\mathcal{AS}$ = $(\mathcal{T},\mathcal{A},\mathcal{N})$
constructed in Example 1.
Since $max\{ depth(\mathcal{T}), depth(\mathcal{A}), depth(\mathcal{N}) \} = 2$,
we let $k$ = 2 and execute the algorithm Adaptation$(\mathcal{AS}, k)$.

By this algorithm we will get two results $(\mathcal{A}_{1}'', \mathcal{R}_{1})$
and $(\mathcal{A}_{2}'', \mathcal{R}_{2})$, where
$\mathcal{R}_{1}$ = $\{ Tamoxifen(y) \}$,
$\mathcal{A}_{1}''$ = $\{ C(Mary)$, $\exists hasGene.CYP2D6(Mary)$,
$\exists TreatBy$. $(Anti\textrm{-}oestrogen$ $\sqcap$ $\exists metabolizedTo$.$(Compounds$ $\sqcap \exists bindto.OestrogenReceptor))$ $(Mary) \}$,
$\mathcal{R}_{2} = \{ TreatBy(Mary,y) \}$,
and $\mathcal{A}_{2}''$ = $\{ C(Mary)$, $\exists hasGene.CYP2D6$ $(Mary) \}$.
Since $\mathcal{R}_{2}$ contains a role assertion $TreatBy(Mary,y)$ which indicates that
all the information related to $y$ is lost in $\mathcal{A}_{2}''$,
the user should select $(\mathcal{A}_{1}'', \mathcal{R}_{1})$
and get a solution described by $\mathcal{A}_{1}''$.
\qed
\end{example}

\section{Discussion and Related Work} \label{section:related work}

The idea of applying KB revision theory to adaptation in CBR was proposed by Lieber \cite{lieber:07}.
Based on a classical revision operator in propositional logic,
a framework for adaptation was presented and
it was demonstrated that the adaptation process should satisfy the AGM postulates.
This idea was extended by Cojan and Lieber \cite{cojan:10} to deal with adaptation based on the DL $\mathcal{ALC}$.
Based on an extension of the classical tableau method used for deductive inferences in $\mathcal{ALC}$,
an algorithm for adapting cases represented in $\mathcal{ALC}$ was proposed.
It was shown that,
except for the requirements on syntax-independence and minimality of change (i.e., $\textbf{R4}$ and $\textbf{R5}$ in our paper),
all the other requirements specified by the AGM postulates (i.e., $\textbf{R1}$-$\textbf{R3}$ in our paper) are satisfied by their algorithm.



From the point of view of KB revision in DLs, it is a great challenge to design revision operators or algorithms
that satisfy the requirements specified by the AGM postulates \cite{flouris:05}.
In the literature, there are two kinds of approaches,
i.e., MBAs \cite{kharlamov:13} and FBAs \cite{calvanese:10,lenzerini:11,wiener:06}, for the instance-level KB revision problem in DLs.
As we analyzed in Section \ref{section:Different Approaches},
they either do not satisfy the requirements specified by $\textbf{R4}$ and $\textbf{R5}$,
or only work well for DLs of the \emph{DL-Lite} family.

Our method can be viewed as a combination of MBAs and FBAs.
On the one hand, in our algorithm,
the revision graph $\mathcal{G}$ constructed by the procedure B-MW($\langle \mathcal{T},\mathcal{A} \rangle, k$)
can be seen as a non-redundant, $k$-depth-bounded model for the KB $\langle \mathcal{T},\mathcal{A} \rangle$,
and therefore our revision process essentially works on models.
On the other hand, our revision process makes use of $(\mathcal{A}_{\mathcal{G}},\mathcal{N})$-repairs
which inherits some ideas of FMAs based on justifications.
As a result, our algorithm not only satisfies the requirements $\textbf{R4}$ and $\textbf{R5}$,
but also works for the DL $\mathcal{EL_{\bot}}$.

For the adaptation setting $\mathcal{AS}$ = $(\mathcal{T},\mathcal{A},\mathcal{N})$ defined in this paper,
$\mathcal{A}$ only contains knowledge on the solution of the selected source case.
We can define $\mathcal{A}$ to contain knowledge on the problem description of the source case also,
and the algorithm still works for such a new definition.
%

Given an adaptation setting, our algorithm will return a finite number of pairs $(\mathcal{A}_{i}'', \mathcal{R}_{i})$ $(1 \leq i \leq n)$,
and it is left to the user to select the best solution according to the sets $\mathcal{R}_{i}$.
We can extend the algorithm to recommend solutions automatically.
For example, if some $\mathcal{R}_{i}$ contains role assertions, then the corresponding solution $\mathcal{A}_{i}''$
will be the last one to be considered.
Furthermore, we can define a selection function according to the user's selection criteria,
and enable our algorithm to return only one best solution.


\section{Conclusion and Future Work} \label{section:conclusion}

We studied the adaptation problem of CBR in the DL $\mathcal{EL_{\bot}}$.
A formalism for adaptation based on $\mathcal{EL_{\bot}}$ was presented,
and in this formalism the adaptation task was modeled as the instance-level KB revision problem in $\mathcal{EL_{\bot}}$.
We have illustrated that existing revision operators and algorithms in DLs did not
work for the adaptation setting based on $\mathcal{EL_{\bot}}$.
By combining MBAs and FBAs, we presented a new algorithm for the adaptation problem,
and showed that our algorithm behaves well for $\mathcal{EL_{\bot}}$
in that it satisfies the requirements proposed in the literature for revision operators.


For future work, we will extend our method to support adaptation based on $\mathcal{EL^{++}}$ \cite{baader:05b}.
Another work is to implement and optimize our algorithm
and test its feasibility in practice.

\section*{Appendix. Proofs for Propositions and Theorems}

We introduce a lemma for proving Proposition 1.

\begin{lemma} \label{theorem:normal witness for k}
Given a KB $\mathcal{K} = \langle \mathcal{T},\mathcal{A} \rangle$
and a non-negative integer $k$,
let $\mathcal{G}$ = $(V,E,\mathcal{L})$
be the revision graph generated after executing Step 2 of the procedure
B-MW($\mathcal {K}, k$),
and let $\mathcal{A}_{\mathcal{G}}$ be the ABox representation of $\mathcal{G}$.
Then, for any ABox assertion $\alpha$ with $sig(\alpha)$ $\subseteq$ $sig(\mathcal{K})$
and $depth(\alpha)$ $\leq$ $k$,
$\mathcal{K}$ $\models$ $\alpha$  iff  $\langle \emptyset,\mathcal{A}_{\mathcal{G}} \rangle$ $\models$ $\alpha$.
\end{lemma}
\begin{proof}
(\textbf{The If direction})
Suppose $\langle \emptyset,\mathcal{A}_{\mathcal{G}} \rangle$ $\models$ $\alpha$.
It is obvious that $\langle \mathcal{T},\mathcal{A}_{\mathcal{G}} \rangle$ $\models$ $\alpha$.
We need to show that
$\langle \mathcal{T},\mathcal{A} \rangle$ $\models$ $\alpha$.

If $\mathcal{A}$ is inconsistent w.r.t. $\mathcal{T}$, then the result is trivial.

Now suppose $\mathcal{A}$ is consistent w.r.t. $\mathcal{T}$.
Let $I=(\triangle^{I}, \cdot^{I})$ be any interpretation with $I \models$ $\mathcal{T}$ and $I \models$ $\mathcal{A}$.
Let $\mathcal{G}_{0}$ be the revision graph constructed in Step 1 of the procedure B-MW($\mathcal {K}, k$).
Let $n$ be the number of times of applying expansion rules in Step 2, and
let $\mathcal{G}_{k}$ $(1 \leq k \leq n)$ be the revision graph after the $i$-th expansion.
Let $\mathcal{A}_{k}$ $(0 \leq k \leq n)$ be the ABox representation of $\mathcal{G}_{i}$.
Then we have $\mathcal{A}_{0}$ = $\mathcal{A}$ and $\mathcal{A}_{n}$ = $\mathcal{A}_{\mathcal{G}}$.

Firstly, by induction on $k$ $(0 \leq k \leq n)$, we show that
for each $\mathcal{A}_{k}$
there exists some interpretation $I_{k}=(\triangle^{I_{k}}, \cdot^{I_{k}})$ with
\begin{itemize}
  \item $I_{k} \models$ $\mathcal{T}$,
  \item $I_{k} \models$ $\mathcal{A}_{k}$,
  \item $\triangle^{I_{k}}$ = $\triangle^{I}$, and
  \item $X^{I_{k}}$ = $X^{I}$ for any $X \in sig(\mathcal{K})$.
\end{itemize}

(\emph{Base case.})
If $k=0$, then we can construct $I_{0}$ = $I$ and get the interpretation we want.

(\emph{Inductive step.})
Let $I_{k}=(\triangle^{I_{k}}, \cdot^{I_{k}})$ be an interpretation satisfying the above property.
There are two cases to be investigated.

(\emph{Case 1.})
Suppose the revision graph $\mathcal{G}_{k+1}$ is generated by applying
the GCI$_{Ind}$-rule, GCI$_{Var}$-rule or $\sqcap$-rule
on the graph $\mathcal{G}_{k}$. Then it is obvious that
$I_{k} \models$ $\mathcal{A}_{k+1}$.
Construct $I_{k+1}$ = $I_{k}$. Then $I_{k+1}$ is the interpretation we want.

(\emph{Case 2.})
Suppose $\mathcal{G}_{k+1}$ is generated by applying
the $\exists$-rule for some node $x$ in $\mathcal{G}_{k}$
and some concept $\exists R.C$ $\in$ $\mathcal{L}(x)$.
Let $z$ be the new variable introduced by applying this rule.
Since $I_{k} \models$ $\mathcal{A}_{k}$, we have $I_{k} \models$ $(\exists R.C)(x)$.
Therefore, there exists some $p \in$ $\triangle^{I_{k}}$ with
$\langle x^{I_{k}},p \rangle \in R^{I_{k}}$ and $p \in C^{I_{k}}$. Construct an
interpretation $I_{k+1}=(\triangle^{I_{k+1}}, \cdot^{I_{k+1}})$ as follows:
\begin{itemize}
  \item $\triangle^{I_{k+1}}$ = $\triangle^{I_{k}}$,
  \item $X^{I_{k+1}}$ = $X^{I_{k}}$ for any $X \in sig(\mathcal{T}) \cup sig(\mathcal{A}_{k})$, and
  \item $z^{I_{k+1}}$ = $p$.
\end{itemize}
It is obvious that $I_{k+1} \models$ $\mathcal{A}_{k+1}$ and $I_{k+1}$ is the interpretation we want.

Secondly, from $I_{n} \models$ $\mathcal{A}_{n}$,
$I_{n} \models$ $\mathcal{T}$,
$\mathcal{A}_{n}$ = $\mathcal{A}_{\mathcal{G}}$ and
$\langle \mathcal{T},\mathcal{A}_{\mathcal{G}} \rangle$ $\models$ $\alpha$, we have
$I_{n} \models$ $\alpha$.
Furthermore, since $sig(\alpha)$ $\subseteq$ $sig(\mathcal{K})$ and $X^{I_{n}}$ = $X^{I}$ for any $X \in sig(\mathcal{K})$,
we have $I \models$ $\alpha$.
Therefore, we have $\langle \mathcal{T},\mathcal{A} \rangle$ $\models$ $\alpha$.

(\textbf{The Only-if direction})
Suppose $\langle \mathcal{T},\mathcal{A} \rangle$ $\models$ $\alpha$.
We need to show that $\langle \emptyset, \mathcal{A}_{\mathcal{G}} \rangle$ $\models$ $\alpha$.

If $\mathcal {A}_{\mathcal{G}}$ is
inconsistent, then the result is trivial.

Now suppose $\mathcal {A}_{\mathcal{G}}$ is consistent.
Let $I=(\triangle^{I}, \cdot^{I})$ be any interpretation with
$I \models$ $\mathcal{A}_{\mathcal{G}}$.
Construct a revision graph $\mathcal{G}'$ = $(V',E',\mathcal{L}')$ as follows:
\begin{enumerate}
  \item Initialize $\mathcal{G}'$ = $\mathcal{G}$.
  \item Expand $\mathcal{G}'$ by applying the following rules, until none of these rules is applicable:
      \begin{itemize}
          \item if $C \sqsubseteq D$ $\in$ $\mathcal{T}$, $D \notin \mathcal{L}'(x)$,
                $x$ $\notin$ $N_{I}^{\mathcal {K}}$, and
                $\langle \mathcal{T}, \{ E(x) \; | \; E \in \mathcal{L}'(x) \} \rangle$ $\models$ $C(x)$,
                then set $\mathcal{L}'(x) = \mathcal{L}'(x) \cup \{ D \}$;
          \item if $C_{1} \sqcap C_{2}$ $\in$ $\mathcal{L}'(x)$, and $\{ C_{1}, C_{2} \} \nsubseteq \mathcal{L}'(x)$,
                then set $\mathcal{L}'(x) = \mathcal{L}'(x) \cup \{ C_{1}, C_{2} \}$;
          \item if $\exists R.C$ $\in$ $\mathcal{L}'(x)$, $x$ has no successor $z$ with $C \in \mathcal{L}'(z)$,
                then introduce a new variable $z$, set $V' = V' \cup \{ z \}$, $E' = E' \cup \{ \langle x,z \rangle  \}$,
                $\mathcal{L}'(z) =$  $\{ C \}$, and $\mathcal{L}'(\langle x,z \rangle ) = \{ R \}$.
      \end{itemize}
\end{enumerate}
Let $\mathcal{A}_{\mathcal{G}'}$ be the ABox representation of $\mathcal{G}'$.

Based on $I$ and $\mathcal{G}'$, construct an interpretation $I'=(\triangle^{I'}, \cdot^{I'})$ as follows:
\begin{enumerate}
  \item Initialize $I'$ = $I$.
  \item For every $x \in V' \setminus V$, expand $I'$ as follows:
      \begin{itemize}
        \item set $\triangle^{I'}$ = $\triangle^{I'}$ $\cup$ $\{ x \}$;
        \item set $x^{I'}$ = $x$;
        \item let $C_{1}$, ..., $C_{m}$ be all the concept names contained in $\mathcal{L}'(x)$,
              then set $C_{i}^{I'}$ = $C_{i}^{I'}$ $\cup$ $\{ x \}$ for each $1 \leq i \leq m$;
        \item let $y$ be the predecessor of $x$ in graph $\mathcal{G}'$,
              and let $R$ be the concept name contained in $\mathcal{L}'(\langle y,x \rangle )$,
              then set $R^{I'}$ = $R^{I'}$ $\cup$ $\{ \langle y,x \rangle \}$.
      \end{itemize}
\end{enumerate}

According to the above constructions, it is obvious that $I'$ $\models$ $\mathcal{A}_{\mathcal{G}'}$
and $I'$ $\models$ $\mathcal{T}$. At the same time, we have $\mathcal{A}$ $\subseteq$
$\mathcal{A}_{\mathcal{G}}$ $\subseteq$ $\mathcal{A}_{\mathcal{G}'}$
and consequently $I'$ $\models$ $\mathcal{A}$.
Therefore, from $\langle \mathcal{T},\mathcal{A} \rangle$ $\models$ $\alpha$,
we can get $I'$ $\models$ $\alpha$.

By investigating the relationship between $I$ and $I'$, we have the following results:
\begin{itemize}
  \item $a^{I}$ = $a^{I'}$ for every individual name $a$ $\in$ $sig(\mathcal{K})$;
  \item for every concept name $C$ $\in$ $sig(\mathcal{K})$, $x$ $\in$
        $C^{I'} \setminus C^{I}$ only if $level(x) > k$, and
  \item for every role name $R$ $\in$ $sig(\mathcal{K})$, $\langle x,y \rangle$ $\in$
        $R^{I'} \setminus R^{I}$ only if $level(y) > k$.
\end{itemize}
Since $sig(\alpha)$ $\subseteq$ $sig(\mathcal{K})$
and $depth(\alpha)$ $\leq$ $k$,
from $I'$ $\models$ $\alpha$ we have $I$ $\models$ $\alpha$.

Therefore, we have $\langle \emptyset, \mathcal{A}_{\mathcal{G}} \rangle$ $\models$ $\alpha$.
\qed
\end{proof}

\vspace{10pt}
\noindent
\textbf{Proof of Proposition 1.}
Let $\mathcal{G}_{0}$ = $(V_{0},E_{0},\mathcal{L}_{0})$
be the revision graph generated after executing Step 2 of the procedure B-MW($\mathcal {K}, k$),
and let $\mathcal{A}_{\mathcal{G}_{0}}$ be the ABox representation of $\mathcal{G}_{0}$.
By Lemma 1, we have $\mathcal{K}$ $\models$ $\alpha$  iff  $\langle \emptyset, \mathcal{A}_{\mathcal{G}_{0}} \rangle$ $\models$ $\alpha$.

Let $\mathcal{G}_{1}$ = $(V_{1},E_{1},\mathcal{L}_{1})$ be the graph generated by Step 3 of the procedure, and
let $\mathcal{A}_{\mathcal{G}_{1}}$ be the ABox representation of
$\mathcal{G}_{1}$.
Based on the expansion rules in Fig. 1,
it is obvious that
$\langle \emptyset, \mathcal{A}_{\mathcal{G}_{0}} \rangle$ $\models$ $\alpha$ iff
$\langle \emptyset, \mathcal{A}_{\mathcal{G}_{1}} \rangle$ $\models$ $\alpha$.


Let $\mathcal{G}_{2}$ = $(V_{2},E_{2},\mathcal{L}_{2})$, ...,  $\mathcal{G}_{m}$ = $(V_{m},E_{m},\mathcal{L}_{m})$
be the sequence of all graphs generated during the execution of Step 4
of the procedure B-MW($\mathcal {K}, k$).
Let $\mathcal{G}_{i}$ = $(V_{i},E_{i},\mathcal{L}_{i})$ $(2 \leq i \leq m)$ is generated by
dropping a redundant branch
$\mathcal{B}_{i} = (V_{\mathcal{B}_{i}},E_{\mathcal{B}_{i}},\mathcal{L}_{\mathcal{B}_{i}})$ for which the root node is $x_{i}$.
Let $\mathcal{A}_{\mathcal{G}_{i}}$ $(2 \leq i \leq m)$ be the ABox representation of
$\mathcal{G}_{i}$.
Then we have $\mathcal{G}_{m}$ = $\mathcal{G}$
and $\mathcal {A}_{\mathcal{G}_{m}}$ = $\mathcal {A}_{\mathcal{G}}$.

For any concept assertion $C(x_{i})$ with $rd(C)$ $\leq$ $k-level(x_{i})$,
it is obvious that $\mathcal{A}_{\mathcal{G}_{i}}$ $\models$ $C(x_{i})$  iff
$\mathcal{A}_{\mathcal{G}_{i-1}}$ $\models$ $C(x_{i})$. So,
we have
$\langle \emptyset, \mathcal{A}_{\mathcal{G}_{i-1}} \rangle$ $\models$ $\alpha$ iff
$\langle \emptyset, \mathcal{A}_{\mathcal{G}_{i}} \rangle$ $\models$ $\alpha$.

To sum up, we have
$\mathcal{K}$ $\models$ $\alpha$  iff  $\langle \emptyset, \mathcal{A}_{\mathcal{G}} \rangle$ $\models$ $\alpha$.
\qed

\vspace{10pt}

We introduce two lemmas for proving Theorem 1 and Theorem 2.

\begin{lemma} \label{theorem:graph equivalent}
Let $\mathcal{K}= \langle \mathcal{T},\mathcal{A} \rangle$ and
$\mathcal{K}'= \langle \mathcal{T},\mathcal{A}' \rangle$ be two consistent KBs,
let $k$ be an integer with $k$ $\geq$ $depth(\mathcal{K})$ and $k$ $\geq$ $depth(\mathcal{K}')$.
Let $\mathcal{G}$ = $(V,E,\mathcal{L})$ and $\mathcal{G}'$ = $(V',E',\mathcal{L}')$
be the revision graphs returned by B-MW($\mathcal{K}, k$) and B-MW($\mathcal{K}', k$) respectively.
If $\mathcal{K}$ $\equiv$ $\mathcal{K}'$,
then there exists a bijection $f: V \rightarrow V'$ such that
\begin{itemize}
  \item $f(x)$ $=$ $x$ if $x$ is an individual name,
  \item $\mathcal{L}(x)$ $=$ $\mathcal{L}'(f(x))$ for every node $x \in V$,
  \item $\langle x,y \rangle$ $\in E$ iff $\langle f(x),f(y) \rangle$ $\in$ $E'$
         for every pair of nodes $x,y$, and
  \item $\mathcal{L}(\langle x,y \rangle)$ $=$ $\mathcal{L}'(\langle f(x),f(y) \rangle)$
        for every edge $\langle x,y \rangle$ $\in E$.
\end{itemize}
\end{lemma}
\begin{proof}
We prove the result by constructing the function $f: V \rightarrow V'$.
Let $\mathcal{A}_{\mathcal{G}}$, $\mathcal{A}_{\mathcal{G}'}$ be the ABox representation of
$\mathcal{G}$ and $\mathcal{G}'$ respectively.
Then $\mathcal{A}_{\mathcal{G}}$ and $\mathcal{A}_{\mathcal{G}'}$
only contains of role assertions and concept assertions of the form $C(x)$
with $C$ $\in$ $N_{C}^{\mathcal{K}}$.
Furthermore, for every individual name $x$, we have $x \in V$ iff $x \in V'$.

Firstly, set $f(x) = x$ for every individual name $x$ in $V$.

Let $a$ be any individual name in $V$,
and let $C$ be any concept name in $\mathcal{L}(a)$.
By the construction of $\mathcal{A}_{\mathcal{G}}$,
we have $C(a)$ $\in$ $\mathcal{A}_{\mathcal{G}}$ and consequently
$\langle \emptyset,\mathcal{A}_{\mathcal{G}} \rangle$ $\models$ $C(a)$.
By Proposition \ref{theorem:relationship between ABox and MNW},
we have $\langle \mathcal{T},\mathcal{A} \rangle$ $\models$ $C(a)$.
Since $\mathcal{K}$ $\equiv$ $\mathcal{K}'$,
we have $\langle \mathcal{T},\mathcal{A}' \rangle$ $\models$ $C(a)$.
By Proposition \ref{theorem:relationship between ABox and MNW} again,
we have $\langle \emptyset,\mathcal{A}_{\mathcal{G}'} \rangle$ $\models$ $C(a)$
and consequently $C(a)$ $\in$ $\mathcal{A}_{\mathcal{G}'}$.
By the construction of $\mathcal{A}_{\mathcal{G}'}$,
we have $C$ $\in$ $\mathcal{L}'(a)$.
Therefore, we have $\mathcal{L}(a)$ $\subseteq$ $\mathcal{L}'(a)$,
i.e., $\mathcal{L}(a)$ $\subseteq$ $\mathcal{L}'(f(a))$.
Similarly, we can get $\mathcal{L}'(f(a))$ $\subseteq$ $\mathcal{L}(a)$.

Let $\langle a,b \rangle$ be any edge in $E$ with $a,b$ individual names,
and let $r$ be any role name in $\mathcal{L}(\langle a,b \rangle)$.
By the construction of $\mathcal{A}_{\mathcal{G}}$,
we have $r(a,b)$ $\in$ $\mathcal{A}_{\mathcal{G}}$ and consequently
$\langle \emptyset,\mathcal{A}_{\mathcal{G}} \rangle$ $\models$ $r(a,b)$.
By Proposition \ref{theorem:relationship between ABox and MNW},
we have $\langle \mathcal{T},\mathcal{A} \rangle$ $\models$ $r(a,b)$.
Since $\mathcal{K}$ $\equiv$ $\mathcal{K}'$,
we have $\langle \mathcal{T},\mathcal{A}' \rangle$ $\models$ $r(a,b)$.
By Proposition \ref{theorem:relationship between ABox and MNW} again,
we have $\langle \emptyset,\mathcal{A}_{\mathcal{G}'} \rangle$ $\models$ $r(a,b)$
and consequently $r(a,b)$ $\in$ $\mathcal{A}_{\mathcal{G}'}$.
By the construction of $\mathcal{A}_{\mathcal{G}'}$,
we have $\langle a,b \rangle$ $\in$ $E'$
and $r$ $\in$ $\mathcal{L}'(\langle a,b \rangle)$.
Therefore, we have
$\mathcal{L}(\langle a,b \rangle)$ $\subseteq$ $\mathcal{L}'(\langle f(a),f(b) \rangle)$.
Similarly, we can get $\mathcal{L}'(\langle f(a),f(b) \rangle)$ $\subseteq$ $\mathcal{L}(\langle a,b \rangle)$.

To sum up, we have $\mathcal{L}(a)$ $=$ $\mathcal{L}'(f(a))$ for every individual name $a$.
Furthermore, for every pair of individual names $a$ and $b$,
we have $\langle a,b \rangle$ $\in E$ iff $\langle f(a),f(b) \rangle$ $\in$ $E'$, and
$\mathcal{L}(\langle a,b \rangle)$ $=$ $\mathcal{L}'(\langle f(a),f(b) \rangle)$.

Secondly, let $a_{1}$, ..., $a_{m}$ be all the individual names contained in $V$ and $V'$.
Let $V_{i}$ $\subseteq$ $V$ $(1 \leq i \leq m)$ be a set composed of $a_{i}$ and all the
variables which are descendants of $a_{i}$ in $\mathcal{G}$, and
let $V_{i}'$ $\subseteq$ $V'$ $(1 \leq i \leq m)$ be a set composed of $a_{i}$ and all the
variables which are descendants of $a_{i}$ in $\mathcal{G}'$.
Let $\mathcal{G}_{i}$ $(1 \leq i \leq m)$ be the subgraph induced on $\mathcal{G}$
by the vertex subset $V_{i}$, and
let $\mathcal{G}_{i}'$ $(1 \leq i \leq m)$ be the subgraph induced on $\mathcal{G}'$
by the vertex subset $V_{i}'$.
For any node $x$ of $\mathcal{G}_{i}$ $(1 \leq i \leq m)$,
let $Roll(x)$ be a concept constructed inductively as follows:
\begin{itemize}
  \item if $x$ has no successor, then
       $Roll(x)$ $=$ $\mathop{\bigsqcap}\limits_{C \in \mathcal{L}(x)} C $, else
  \item $Roll(x)$ $=$ $\exists r_{1}.Roll(y_{1})$ $\sqcap$ ... $\sqcap$
       $\exists r_{t}.Roll(y_{t})$ $\sqcap$ $\mathop{\bigsqcap}\limits_{C \in \mathcal{L}(x)} C $,
       where $y_{1}, ..., y_{t}$ are all the successors of $x$ in $\mathcal{G}$,
       and $r_{i}$ $(1 \leq i \leq t)$ is the role name contained in $\mathcal{L}(\langle x,y_{i} \rangle)$.
\end{itemize}
Similarly, for any node $x$ of $\mathcal{G}_{i}'$ $(1 \leq i \leq m)$,
let $Roll'(x)$ be a concept constructed inductively as follows:
\begin{itemize}
  \item if $x$ has no successor, then
       $Roll'(x)$ $=$ $\mathop{\bigsqcap}\limits_{C \in \mathcal{L}'(x)} C $, else
  \item $Roll'(x)$ $=$ $\exists r_{1}.Roll'(y_{1})$ $\sqcap$ ... $\sqcap$
       $\exists r_{t}.Roll'(y_{t})$ $\sqcap$ $\mathop{\bigsqcap}\limits_{C \in \mathcal{L}'(x)} C $,
       where $y_{1}, ..., y_{t}$ are all the successors of $x$ in $\mathcal{G}'$,
       and $r_{i}$ $(1 \leq i \leq t)$ is the role name contained in $\mathcal{L}'(\langle x,y_{i} \rangle)$.
\end{itemize}

For each individual name $a_{i}$ $(1 \leq i \leq m)$,
let $Roll(a_{i})$ be of the form $\exists r_{1}.Roll(y_{1})$ $\sqcap$ ... $\sqcap$
$\exists r_{t}.Roll(y_{t})$ $\sqcap$ $\mathop{\bigsqcap}\limits_{C \in \mathcal{L}(a_{i})} C$,
and let $Roll'(a_{i})$ be of the form $\exists r_{1}'.Roll'(y_{1}')$ $\sqcap$ ... $\sqcap$
$\exists r_{s}'.Roll'(y_{s}')$ $\sqcap$ $\mathop{\bigsqcap}\limits_{C \in \mathcal{L}'(a_{i})} C$.
Then we have $\langle \emptyset,\mathcal{A}_{\mathcal{G}} \rangle$ $\models$ $Roll(a_{i})(a_{i})$
and $\langle \emptyset,\mathcal{A}_{\mathcal{G}'} \rangle$ $\models$ $Roll'(a_{i})(a_{i})$.
By Proposition \ref{theorem:relationship between ABox and MNW},
we have $\langle \mathcal{T},\mathcal{A} \rangle$ $\models$ $Roll(a_{i})(a_{i})$
and $\langle \mathcal{T},\mathcal{A}' \rangle$ $\models$ $Roll'(a_{i})(a_{i})$.
Since $\mathcal{K}$ $\equiv$ $\mathcal{K}'$,
we have $\langle \mathcal{T},\mathcal{A}' \rangle$ $\models$ $Roll(a_{i})(a_{i})$ and
$\langle \mathcal{T},\mathcal{A} \rangle$ $\models$ $Roll'(a_{i})(a_{i})$.
By Proposition \ref{theorem:relationship between ABox and MNW} again,
we have $\langle \emptyset,\mathcal{A}_{\mathcal{G}'} \rangle$ $\models$ $Roll(a_{i})(a_{i})$ and
$\langle \emptyset,\mathcal{A}_{\mathcal{G}} \rangle$ $\models$ $Roll'(a_{i})(a_{i})$.
Therefore, it must be $t=s$.
Furthermore, there must be a one-to-one correspondence from $\{y_{1}, ..., y_{t}\}$
to $\{y_{1}', ..., y_{s}'\}$
(w.l.o.g. we assume $y_{i}$ is corresponding to $y_{i}'$)
such that
$\langle \emptyset,\mathcal{A}_{\mathcal{G}'} \rangle$ $\models$ $Roll(y_{k})(y_{k}')$,
$\langle \emptyset,\mathcal{A}_{\mathcal{G}} \rangle$ $\models$ $Roll'(y_{k}')(y_{k})$,
$r_{k}$ = $r'_{k}$, and $\mathcal{L}(y_{k})$ = $\mathcal{L'}(y_{k}')$
for every $1 \leq k \leq t$.
So, we can continue the construction of the function $f$,
and set $f(y_{k}) = y_{k}'$ for each node $y_{k}$ $(1 \leq k \leq t)$.

Now, for each node $y_{k}$ $(1 \leq k \leq t)$,
let $Roll(y_{k})$ be of the form $\exists r_{k,1}.Roll(y_{k,1})$ $\sqcap$ ... $\sqcap$
$\exists r_{k,u}.Roll(y_{k,u})$ $\sqcap$ $\mathop{\bigsqcap}\limits_{C \in \mathcal{L}(y_{k})} C$, and
let $Roll'(y_{k}')$ be of the form $\exists r_{k,1}'.$ $Roll'(y_{k,1}')$ $\sqcap$ ... $\sqcap$
$\exists r_{k,v}'.Roll'(y_{k,v}')$ $\sqcap$ $\mathop{\bigsqcap}\limits_{C \in \mathcal{L}'(y_{k}')} C$.
From $\langle \emptyset,\mathcal{A}_{\mathcal{G}'} \rangle$ $\models$ $Roll(y_{k})(y_{k}')$ and
$\langle \emptyset,\mathcal{A}_{\mathcal{G}} \rangle$ $\models$ $Roll'(y_{k}')(y_{k})$,
it must be $u=v$.
Furthermore, there must be a one-to-one correspondence from $\{y_{k,1}, ..., y_{k,u}\}$
to $\{y_{k,1}', ..., y_{k,v}'\}$
(w.l.o.g. we assume $y_{k,i}$ is corresponding to $y_{k,i}'$)
such that
$\langle \emptyset,\mathcal{A}_{\mathcal{G}'} \rangle$ $\models$ $Roll(y_{k,j})(y_{k,j}')$,
$\langle \emptyset,\mathcal{A}_{\mathcal{G}} \rangle$ $\models$ $Roll'(y_{k,j}')(y_{k,j})$,
$r_{k,j}$ = $r_{k,j}'$, and $\mathcal{L}(y_{k,j})$ = $\mathcal{L'}(y_{k,j}')$
for every $1 \leq j \leq u$.
So, we can continue the construction of the function $f$,
and set $f(y_{k,j}) = y_{k,j}'$ for each node $y_{k,j}$ $(1 \leq j \leq u)$.

Repeat the above process, we can finish the construction of the
function $f: V \rightarrow V'$ which satisfies the property stated by the lemma.
\qed
\end{proof}

\begin{lemma} \label{theorem:roll up}
For any ABox $\mathcal{A}_{\mathcal{G}}$ and
TBox $\mathcal{T}$, let $\mathcal{A}$ be the ABox returned by
the procedure Rolling($\mathcal{A}_{\mathcal{G}}$, $\mathcal{T}$).
Then, $\langle \mathcal{T},\mathcal{A}_{\mathcal{G}} \rangle$ $\models$ $\mathcal{A}$.
\end{lemma}
\begin{proof}
Let $\mathcal{G}$ = $(V,E,\mathcal{L})$ be the revision-graph representation of $\mathcal{A}_{\mathcal{G}}$.
Let $\mathcal{G}_{0}$ = $(V_{0},E_{0},\mathcal{L}_{0})$ be the revision graph generated by Step 2 of the procedure.
Let $n$ be the number of variables contained in $V_{0}$, and
let $\mathcal{G}_{i}$ = $(V_{i}, E_{i},\mathcal{L}_{i})$ $(1 \leq i \leq n)$
be the revision graph generated after removing $k$ variables from $\mathcal{G}_{0}$ in Step 3.
Let $\mathcal{A}_{i}$ $(0 \leq i \leq n)$ be the ABox representation of $\mathcal{G}_{i}$.
Then we have $\mathcal{A}_{n}$ = $\mathcal{A}$.

If $\mathcal{A}_{\mathcal{G}}$ is inconsistent w.r.t. $\mathcal{T}$, then the result trivial.

Now suppose $\mathcal{A}_{\mathcal{G}}$ is consistent w.r.t. $\mathcal{T}$.
Let $I$ be any interpretation with $I \models$ $\mathcal{T}$ and $I \models$ $\mathcal{A}_{\mathcal{G}}$.
It is obvious that $I \models$ $\mathcal{A}_{0}$.
According to the operations of Step 3, we can get
$I \models$ $\mathcal{A}_{i}$ $(1 \leq i \leq n)$
from $I \models$ $\mathcal{T}$ and $I \models$ $\mathcal{A}_{i-1}$.
Therefore we have $I \models$ $\mathcal{A}_{n}$, i.e., $I \models$ $\mathcal{A}$.
So, we have $\langle \mathcal{T},\mathcal{A}_{\mathcal{G}} \rangle$ $\models$ $\mathcal{A}$.
\qed
\end{proof}

\vspace{10pt}
\noindent
\textbf{Proof of Theorem 1.}
The first two statements are obvious. Here we prove the third one.

Suppose $\mathcal{N}$ is consistent w.r.t. $\mathcal{T}$.
Then the set $S_\mathcal{R}$ of $(\mathcal{A}_{\mathcal{G}},\mathcal{N})$-repairs constructed in Algorithm 1 is not empty.
For any $\mathcal{R}_{i}$ $\in$ $S_\mathcal{R}$, let $\mathcal{A}_{i}$ = $\mathcal{A}_{\mathcal{G}} \setminus \mathcal{R}_{i}$,
then, by the definition of $(\mathcal{A}_{\mathcal{G}},\mathcal{N})$-repairs,
we have that $\mathcal{A}_{i}$ $\cup$ $\mathcal{N}$ is consistent w.r.t. $\mathcal{T}$.
Let $\mathcal{A}_{i}'$ = Rolling($\mathcal{A}_{i}$, $\mathcal{T}$).
By Lemma \ref{theorem:roll up}, We have
$\mathcal{A}_{min}$ $\models_{\mathcal{T}}$ $\mathcal{A}_{0}$.
Therefore, $\mathcal{A}_{i}'$ $\cup$ $\mathcal{N}$ is also consistent w.r.t. $\mathcal{T}$.
So, for each pair $(\mathcal{A}_{i}'', \mathcal{R}_{i})$ returned by the algorithm,
$\mathcal{A}_{i}''$ is consistent w.r.t. $\mathcal{T}$.
\qed

\vspace{13pt}
\noindent
\textbf{Proof of Theorem 2.}
%
Let $\mathcal{G}_{i}$ = B-MW($\langle \mathcal{T},\mathcal{A}_{i} \rangle, k$), and
let $\mathcal{A}_{\mathcal{G}_{i}}$ be the ABox representation of $\mathcal{G}_{i}$ $(i = 1, 2)$.
Since $k \geq$ $max \{ depth(\mathcal{T})$, $depth(\mathcal{A}_{1})$, $depth(\mathcal{N}_{1})$,
$depth(\mathcal{A}_{2})$, $depth(\mathcal{N}_{2}) \}$,
we know that between $\mathcal{G}_{1}$ and $\mathcal{G}_{2}$ there exists an isomorphism relationship
specified by Lemma \ref{theorem:graph equivalent}.
In other words, there exists a substitution $\sigma$ of variables
such that $\sigma(\mathcal{A}_{\mathcal{G}_{1}})$ = $\mathcal{A}_{\mathcal{G}_{2}}$.

Let $\mathcal{J}_{1}$ $\subseteq$ $\mathcal{A}_{\mathcal{G}_{1}}$
be any $(\mathcal{A}_{\mathcal{G}_{1}},\mathcal{N}_{1})$-justification for a clash w.r.t. $\mathcal{T}$, and
let $\mathcal{J}_{2}$ = $\sigma(\mathcal{J}_{1})$.
From $\langle \mathcal{T}, \mathcal{N}_{1} \rangle$ $\equiv$ $\langle \mathcal{T}, \mathcal{N}_{2} \rangle$ and
$\sigma(\mathcal{A}_{\mathcal{G}_{1}})$ = $\mathcal{A}_{\mathcal{G}_{2}}$,
we have that $\mathcal{J}_{2}$ $\subseteq$ $\mathcal{A}_{\mathcal{G}_{2}}$
and $\mathcal{J}_{2}$ is a $(\mathcal{A}_{\mathcal{G}_{2}},\mathcal{N}_{2})$-justification for a clash w.r.t. $\mathcal{T}$.

Let $\mathcal{R}_{2}$ = $\sigma(\mathcal{R}_{1})$.
Since $\mathcal{R}_{1}$ is a $(\mathcal{A}_{\mathcal{G}_{1}},\mathcal{N}_{1})$-repair for clashes w.r.t. $\mathcal{T}$,
from $\langle \mathcal{T}, \mathcal{N}_{1} \rangle$ $\equiv$ $\langle \mathcal{T}, \mathcal{N}_{2} \rangle$ and
$\sigma(\mathcal{A}_{\mathcal{G}_{1}})$ = $\mathcal{A}_{\mathcal{G}_{2}}$,
we have that $\mathcal{R}_{2}$ $\subseteq$ $\mathcal{A}_{\mathcal{G}_{2}}$
and $\mathcal{R}_{2}$ is a $(\mathcal{A}_{\mathcal{G}_{2}},\mathcal{N}_{2})$-repair for clashes w.r.t. $\mathcal{T}$.

Let $\mathcal{A}_{1}$ = $\mathcal{A}_{\mathcal{G}_{1}}$ $\setminus$ $\mathcal{R}_{1}$ and
$\mathcal{A}_{2}$ = $\mathcal{A}_{\mathcal{G}_{2}}$ $\setminus$ $\mathcal{R}_{2}$.
Then we have that $\mathcal{A}_{2}$ = $\sigma(\mathcal{A}_{1})$.
Let $\mathcal{A}_{1}'$ = Rolling($\mathcal{A}_{1}$, $\mathcal{T}$) and
$\mathcal{A}_{2}'$ = Rolling($\mathcal{A}_{2}$, $\mathcal{T}$).
Then we have that $\mathcal{A}_{1}'$ = $\mathcal{A}_{2}'$ and
$\mathcal{A}_{1}''$ = $\mathcal{A}_{1}'$ $\cup$ $\mathcal{N}_{1}$.

Now construct a set $\mathcal{A}_{2}''$ = $\mathcal{A}_{2}'$ $\cup$ $\mathcal{N}_{2}$.
Then the pair $(\mathcal{A}_{2}'', \mathcal{R}_{2})$ will be returned by Adaptation$(\mathcal{AS}_{2}, k)$.
Furthermore, since $\mathcal{A}_{1}'$ = $\mathcal{A}_{2}'$ and
$\langle \mathcal{T}, \mathcal{N}_{1} \rangle$ $\equiv$ $\langle \mathcal{T}, \mathcal{N}_{2} \rangle$,
we have that $\langle \mathcal{T}, \mathcal{A}_{1}'' \rangle$ $\equiv$ $\langle \mathcal{T}, \mathcal{A}_{2}'' \rangle$.
\qed

\vspace{13pt}
\noindent
\textbf{Proof of Theorem 3.}
Let $k$ = $max \{ depth(\mathcal{T}), \mathcal{A}), depth(\mathcal{N}) \}$.
Let $\mathcal{T}'$ be a normal form of $\mathcal{T}$ such that there is no conjunction
in the right hand side of every GCI contained in it, and let $m = |\mathcal{T}'|$.
It is obvious that $m$ is linearly bounded by $|\mathcal{T}|$.

For the revision graph $\mathcal{G}$ returned by the procedure B-MW($\langle \mathcal{T},\mathcal{A} \rangle, k$),
the number of nodes is bounded by $|N_{I}^\mathcal{K}| \times m^{k+1}$.
Therefore, for the ABox representation $\mathcal {A}_{\mathcal{G}}$ of $\mathcal{G}$, $|\mathcal {A}_{\mathcal{G}}|$ is
bounded by $|\mathcal{A}| + $ $|N_{C}^\mathcal{K}| \times |N_{I}^\mathcal{K}| \times m^{k+1}$.
Since every $(\mathcal{A}_{\mathcal{G}},\mathcal{N})$-justification and
every $(\mathcal{A}_{\mathcal{G}},\mathcal{N})$-repair is a subset of $\mathcal {A}_{\mathcal{G}}$,
together with the fact that KB consistency problem is polynomial time in $\mathcal{EL}_{\bot}$,
we can conclude that the time complexity of Algorithm 1
is $O(2^{|N_{C}^\mathcal{K}| \times |N_{I}^\mathcal{K}| \times m^{k+1}})$.
Since $k$ is bounded by some constant,
we get the result stated by the theorem.
\qed


\begin{thebibliography}{6}

\bibitem {alchourron:85}
Alchourr$\acute{o}$n, C.E., G$\ddot{a}$rdenfors, P., Makinson, D.:
On the logic of theory change: Partial meet contraction and
revision functions. J. Symb. Log, 50(2), 510-530 (1985)


\bibitem {baader:03}
Baader, F., Calvanese, D., McGuinness, D., Nardi, D.,
Patel-Schneider, P.F.: The Description
Logic Handbook: Theory, Implementation and Applications.
Cambridge University Press, Cambridge (2003)


\bibitem {baader:05b}
Baader,F., Brandt, S., Lutz, C.: Pushing the $\mathcal{EL}$ envelope.
In: Proc. of the 19th
International Joint Conference on Artificial Intelligence,
pp. 364-369. Morgan Kaufmann (2005)


\bibitem {calvanese:10}
Calvanese, D., Kharlamov, E., Nutt, W., Zheleznyakov, D.:
Evolution of DL-lite knowledge bases.
In: Patel-Schneider, P.F., Pan, Y., Hitzler, P., Mika, P., Zhang, L., Pan, J.Z., Horrocks, I., Glimm, B. (eds.) ISWC 2010.
LNCS, vol. 6496, pp. 112-128.
Springer, Heidelberg (2010)




\bibitem {cojan:10}
Cojan, J., Lieber, J.:
An algorithm for adapting cases represented in an expressive description logic.
In: Bichindaritz, I., Montani, S. (eds.) ICCBR 2010.
LNCS, vol. 6176, pp. 51-65.
Springer, Heidelberg (2010)


\bibitem {consortium:00}   
Consortium, T.G.O.:
Gene Ontology: Tool for the unification of biology.
Nature Genetics, 25, 25-29 (2000)




\bibitem {d'aquin:05}
d'Aquin, M., Lieber, J., Napoli, A.:
Decentralized case-based reasoning for the semantic web.
In: Gil, Y., Motta, E., Benjamins, V.R., Musen, M.A. (eds.) ISWC 2005.
LNCS, vol. 3729, pp. 142-155.
Springer, Heidelberg (2005)


\bibitem {flouris:05}
Flouris, G., Plexousakis, D., Antoniou, G.:
On applying the AGM theory to DLs and OWL.
In: Gil, Y., Motta, E., Benjamins, V.R., Musen, M.A. (eds.) ISWC 2005.
LNCS, vol. 3729, pp. 216-231.
Springer, Heidelberg (2005)


\bibitem {gomez:99}
G$\acute{o}$mez-Albarr$\acute{a}$n, M., Gonz$\acute{a}$lez-Calero, P.A., D$\acute{i}$az-Agudo, B., Fern$\acute{a}$ndez-Conde, C.:
Modelling the CBR life cycle using description logics.
In: Althoff, K.-D., Bergmann, R., Karl Branting, L. (eds.) ICCBR 1999.
LNCS, vol. 1650, pp. 147-161.
Springer, Heidelberg (1999)




\bibitem {horrocks:03}
Horrocks, I., Patel-Schneider, P.F., Harmelen, F.V.: From SHIQ and
RDF to OWL: the making of a web ontology language.
J. Web Semantics, 1(1), 7-26 (2003)


\bibitem {horrocks:07}
Horrocks, I., Sattler, U:
A tableau decision procedure for $\mathcal{SHOIQ}$.
J. Autom. Reasoning, 39(3), 249-276 (2007)




\bibitem {katsuno:91}  
Katsuno, H., Mendelzon, A. O.:
Propositional knowledge base revision and minimal change.
Artificial Intelligence, 52(3), 263-294 (1991)




\bibitem {kharlamov:13}
Kharlamov, E., Zheleznyakov, D., Calvanese, D,:
Capturing model-based ontology evolution at the instance level: The case of DL-Lite.
J. Comput. Syst. Sci., 79(6), 835-872 (2013)




\bibitem {lehmann:12}
Lehmann, K., Turhan, A.Y.:
A Framework for Semantic-Based Similarity Measures for $\mathcal{ELH}$ -Concepts.
In: Cerro, L.F., Herzig, A., Mengin, J. (eds.) JELIA 2012.
LNCS, vol. 7519, pp. 307-319.
Springer, Heidelberg (2012)


\bibitem {lenzerini:11}
Lenzerini, M., Savo, D. F.:
On the evolution of the instance level of DL-Lite knowledge bases.
In: Proc. of the 24th International Workshop on Description Logics
(2011)


\bibitem {lieber:07}
Lieber, J.:
Application of the revision theory to adaptation in case-based reasoning: the conservative adaptation.
In: Weber, R.O.,Richter, M.M. (eds.) ICCBR 2007.
LNCS, vol. 4626, pp. 239-253.
Springer, Heidelberg (2007)


\bibitem {salotti:98}
Salotti, S., Ventos, V.:
Study and formalization of a case-based reasoning system using a description logic.
In: Smyth, B., Cunningham, P. (eds.) EWCBR 1998.
LNCS, vol. 1488, pp. 286-297.
Springer, Heidelberg (1998)




\bibitem {sanchez:11}
S$\acute{a}$nchez-Ruiz-Granados, A.A., Onta$\tilde{n}\acute{o}$n, S., Gonz$\acute{a}$lez-Calero, P.A., Plaza, E.:
Measuring similarity in description logics using refinement operators.
In: Ram, A., Wiratunga, N. (eds.) ICCBR 2011.
LNCS, vol. 6880, pp. 289-303.
Springer, Heidelberg (2011)




\bibitem {spackman:00}  
Spackman, K.:
Managing clinical terminology hierarchies using algorithmic calculation of
subsumption: Experience with SNOMED-RT.
J. American Medical Informatics Assoc.,
Fall Symposium Special Issue (2000)


\bibitem {wiener:06}
Wiener, C.H., Katz, Y., Parsia, B.:
Belief base revision for expressive description logics.
In: Proc. of the 4th International Workshop on OWL: Experiences and Directions (2006)





\end{thebibliography}
\end{document}